%% file: aaai2026.tex
\documentclass[letterpaper]{article} 
\usepackage{aaai2026}  
\usepackage{times}  
\usepackage{helvet}  
\usepackage{courier}  
\usepackage[hyphens]{url}  
\usepackage{graphicx} 
\usepackage{natbib}  
\usepackage{caption} 
\usepackage{algorithm}
\usepackage{algpseudocode}

\usepackage{newfloat}
\usepackage{listings}
\DeclareCaptionStyle{ruled}{labelfont=normalfont,labelsep=colon,strut=off} 
\floatstyle{ruled}
\newfloat{listing}{tb}{lst}{}
\floatname{listing}{Listing}
\usepackage{amssymb}
\usepackage{multirow}
\usepackage{arydshln}
\usepackage{amsmath}
\usepackage{colortbl}

\usepackage{enumitem}
\usepackage{tabularx}
\usepackage{adjustbox}

\usepackage{makecell}
\usepackage{color}
\usepackage{circledtext}
\usepackage{textcomp}
\usepackage{booktabs}

\usepackage{algpseudocode}
\usepackage{caption}
\usepackage{bm}
\usepackage{pifont}

\usepackage{tikz}
\usepackage{subfig}
\usepackage{tkz-kiviat,pgfplots}
\pgfplotsset{compat=newest}
\usepgfplotslibrary{fillbetween}
\usepgfplotslibrary{patchplots} 

\definecolor{myGreen}{RGB}{50, 168, 82}
\definecolor{myYellow}{RGB}{214, 169, 45}
\definecolor{myRed}{RGB}{201, 24, 24}
\definecolor{myBlue}{RGB}{50, 200, 200}
\definecolor{myBrown}{RGB}{140, 50, 50}
\definecolor{myPurple}{RGB}{128, 0, 200} 

\definecolor{ForestGreen}{RGB}{34, 139, 34}
\definecolor{BrickRed}{RGB}{178, 34, 34}

\definecolor{tiffanyblue}{RGB}{129,216,208}
\definecolor{bangdiblue}{RGB}{0,149,182}
\definecolor{kleinblue}{RGB}{0,47,167}

\usepackage[most]{tcolorbox}

\newtcolorbox{promptbox}[2][]{
	width=\linewidth,
	colback=gray!8,
	colframe=gray!8,
	boxsep=0pt,
	left=5pt,
	right=5pt,
	top=2pt,
	bottom=2pt,
        fontupper=\small\linespread{1.2}\selectfont,
	title=#2,#1
}

\newcommand\name{\textit{LangGPS}}

\title{\textit{LangGPS}: Language Separability Guided Data Pre-Selection for Joint Multilingual Instruction Tuning}

\newcounter{corfn}
\newcommand{\corresponding}{%
  \ifnum\value{corfn}=0
    \footnote{Corresponding Authors}%
    \setcounter{corfn}{\value{footnote}}%
  \else
    \footnotemark[\value{corfn}]%
  \fi
}

\author{
    Yangfan Ye\textsuperscript{\rm 1},
    Xiaocheng Feng\textsuperscript{\rm 1,2}\corresponding,
    Xiachong Feng\textsuperscript{\rm 3}\corresponding,
    Lei Huang\textsuperscript{\rm 1},
    Weitao Ma\textsuperscript{\rm 1},
    Qichen Hong\textsuperscript{\rm 4},\\
    Yunfei Lu\textsuperscript{\rm 4},
    Duyu Tang\textsuperscript{\rm 4},
    Dandan Tu\textsuperscript{\rm 4}\corresponding,
    Bing Qin\textsuperscript{\rm 1,2}
}
\affiliations{
    \textsuperscript{\rm 1}Harbin Institute of Technology, 
    \textsuperscript{\rm 2}Peng Cheng Laboratory, 
    \textsuperscript{\rm 3}The University of Hong Kong, 
    \textsuperscript{\rm 4}Huawei Technologies Co., Ltd

    \{yfye,xcfeng,lhuang,wtma,qinb\}@ir.hit.edu.cn, fengxc@hku.hk, \{hongqichen,luyunfei6,tangduyu,tudandan\}@huawei.com
}

\usepackage{bibentry}

\begin{document}

\maketitle
 
\begin{abstract}
Joint multilingual instruction tuning is a widely adopted approach to improve the multilingual instruction-following ability and downstream performance of large language models (LLMs), but the resulting multilingual capability remains highly sensitive to the composition and selection of the training data.
Existing selection methods, often based on features like text quality, diversity, or task relevance, typically overlook the intrinsic linguistic structure of multilingual data.
In this paper, we propose {\name}, a lightweight two-stage pre-selection framework guided by \textit{language separability}—a signal that quantifies how well samples in different languages can be distinguished in the model’s representation space. {\name} first filters training data based on separability scores and then refines the subset using existing selection methods.
Extensive experiments across six benchmarks and 22 languages demonstrate that applying {\name} on top of existing selection methods improves their effectiveness and generalizability in multilingual training, especially for understanding tasks and low-resource languages.
Further analysis reveals that highly separable samples facilitate the formation of clearer language boundaries and support faster adaptation, while low-separability samples tend to function as bridges for cross-lingual alignment.
Besides, we also find that language separability can serve as an effective signal for multilingual curriculum learning, where interleaving samples with diverse separability levels yields stable and generalizable gains.
Together, we hope our work offers a new perspective on data utility in multilingual contexts and support the development of more linguistically informed LLMs.

\textit{- Code: \url{https://github.com/YYF-Tommy/LangGPS}}
\end{abstract}

\section{Introduction}

\begin{figure*}[t]
\centering
\includegraphics[width=0.91\textwidth]{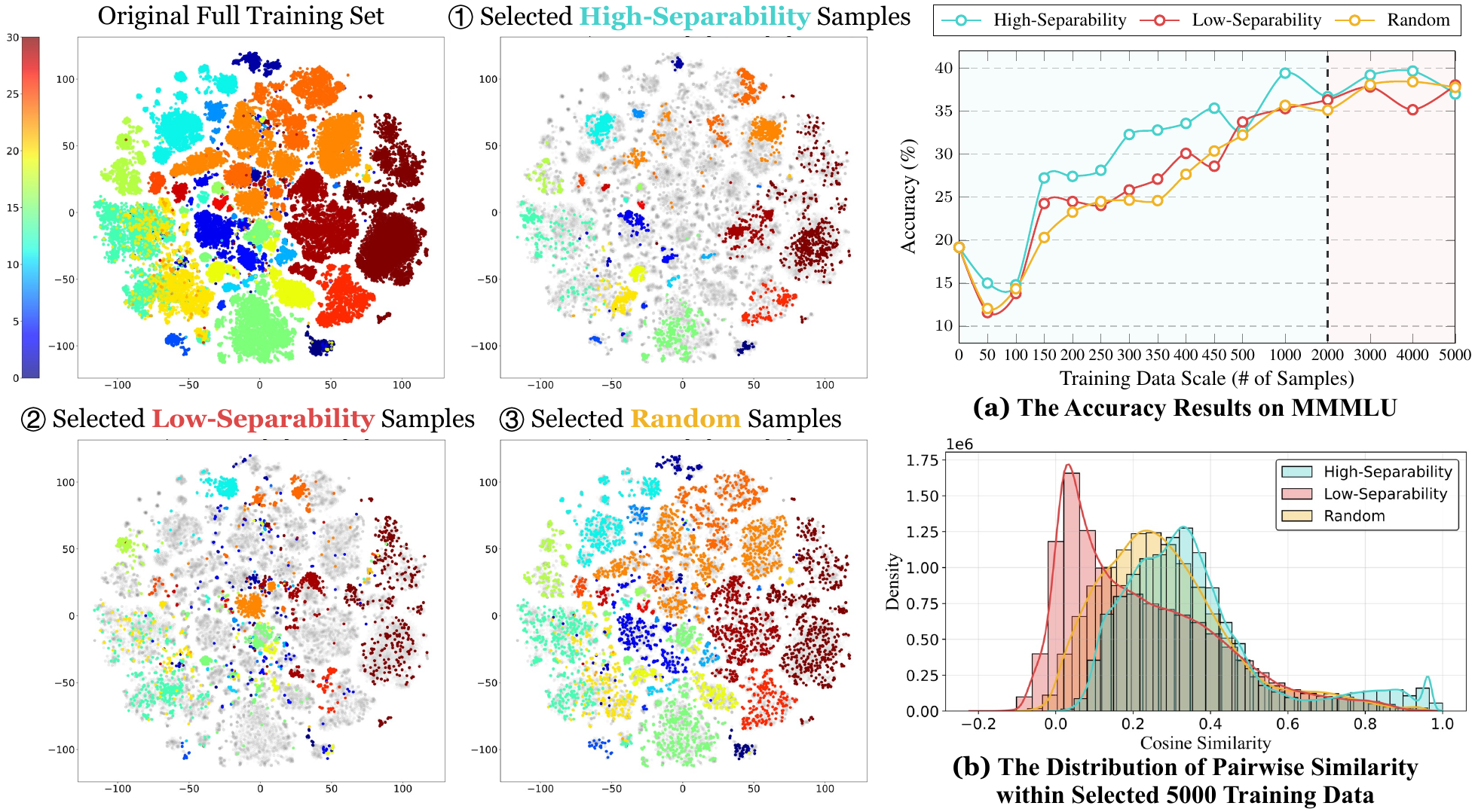}
\caption{The left shows the t-SNE visualization of samples selected by different strategies under same language distribution; each color denotes one of 31 languages. The representations are last-token hidden states from \textit{LLaMA-3.1-8B}. On the right, panel (a) presents the performance of \textit{LLaMA-3.1-8B} after training with the three types of samples on \textit{MMMLU} across training sizes; while panel (b) illustrates the pairwise similarity distributions of selected data. \textbf{(See detailed settings in Appendix~\ref{app:pre_exp})}}
\label{fig:pilot}
\end{figure*}

Large language models (LLMs) have demonstrated remarkable generalization and adaptability across a wide range of downstream tasks in multilingual scenarios~\citep{ye2023language,qin2024multilingual,zhang2023dont,ye2024globesumm,ye2024xtransplant}. A central enabler of such capabilities is joint multilingual instruction tuning, in which LLMs are trained on instruction-following data covering multiple languages~\citep{ouyang2022training,shaham2024multilingual,huo2025enhancing,ye2025cctuning}. This paradigm allows models to understand and follow instructions in multilingual scenarios, fostering more inclusive access to language technologies.

Recent work has emphasized the importance of multilingual instruction datasets that encompass diverse languages, domains, and task formats~\citep{muennighoff2022crosslingual,zhu2023extrapolating,li2023bactrian,singh2024aya}.
These efforts have laid a solid foundation for developing LLMs that can effectively align with user intent across linguistic boundaries.
Despite these advances, the success of multilingual instruction tuning remains highly sensitive to the composition and selection of the training data.
Consequently, \textbf{Data Selection} has emerged as a critical strategy for balancing effectiveness and efficiency in large-scale model training. 
Existing approaches can be typically grouped into two main categories: feature-based methods, which prioritize training samples based on characteristics such as data quality and diversity~\citep{li-etal-2024-quantity, bukharin-etal-2024-data}; and target-dependent methods, which select data according to its estimated relevance to specific target tasks~\citep{xie2023data,xia2024less}. However, little attention has been paid to the intrinsic linguistic structure inherent in multilingual data.

In multilingual scenarios, where the model must learn to follow instructions across diverse and often low-resource languages, we argue that a key prerequisite for effective multilingual instruction following is the model's ability to form and maintain clear linguistic boundaries. Without sufficiently distinguishing between languages, the model risks conflating linguistic patterns, especially in typical low-resource languages.
To formalize this intuition, we introduce the notion of language separability, which quantifies the extent to which the data samples in different languages are distinguishable in the model’s representation space.
Highly separable samples tend to exhibit well-structured, language-specific representations, whereas poorly separable ones often entangle with other languages in the representation space, lacking clear linguistic boundaries.
As shown in Figure~\ref{fig:pilot} (a), our preliminary experiments (see details in Appendix~\ref{app:pre_exp}) on the multilingual MMLU dataset reveal that, under limited training data ($<2000$), training with highly separable samples leads to significantly better performance compared to low-separability or randomly selected samples. However, as Figure~\ref{fig:pilot} (b) illustrates, highly separable samples may also suffer from over-similarity (excessive resemblance among samples), introducing a lack of diversity, which is another critical factor in data selection.

Building on these insights, we propose \textbf{{\name}}, a \textbf{Lang}uage separability \textbf{G}uided data \textbf{P}re-\textbf{S}election strategy for multilingual instruction tuning. 
{\name} adopts a two-stage framework: (1) a lightweight pre-selection stage that filters training data based on estimated language separability, and (2) a fine selection stage that refines the selected subset using existing data selection methods, such as diversity-based, task relevance-based, or gradient-based approaches.
This design integrates scalability, linguistic structure awareness, and full compatibility with existing selection strategies that focus on different aspects, offering a more principled and flexible framework for multilingual instruction tuning.

We conduct extensive experiments across six benchmarks covering both understanding and generation tasks, spanning 22 evaluated languages and two representative LLMs.
Our results find that existing data selection methods often overlook the intrinsic linguistic structure inherent in multilingual training data, leading to inconsistent gains in joint multilingual training.
By leveraging language separability as a guiding signal, {\name} can be seamlessly integrated with existing selection methods and improves their effectiveness and generalizability, especially for understanding tasks and low-resource languages.
Further analysis reveals that highly separable samples are critical for language-specific modeling and fast adaptation, helping models form and main clearer language boundaries, while low-separability samples may serve as valuable bridges for cross-lingual alignment due to their entangled representations across languages.
Moreover, we demonstrate that language separability not only informs data selection, but also enables effective curriculum learning: interleaving training samples with diverse separability levels allows models to absorb both structured and entangled linguistic signals, improving training effectiveness.
These findings highlight the importance of considering linguistic structure in multilingual training pipelines, and we hope they inspire future work on more linguistically aware approaches in multilingual community.

\section{Related Work}

\paragraph{Joint Multilingual Instruction Fine-Tuning.} 
While early work shows that fine-tuning on English-only data can yield cross-lingual generalization~\citep{muennighoff2022crosslingual}, multilingual instruction tuning—i.e., fine-tuning on data from multiple languages—has emerged as a more widely adopted strategy for improving multilingual performance~\citep{ouyang2022training}.
Recent studies often incorporate data augmentation or distillation techniques to enhance multilingual capabilities.
For example, \citet{li2023bactrian} addressed the issue of "translationese" by generating multilingual responses using Google Translate and ChatGPT.
Some other approaches incorporate multilingual examples into English-centric tuning~\citep{shaham2024multilingual}, apply translation-based finetuning to improve semantic alignment~\citep{ranaldi-etal-2024-empowering}, combine translation data, cross-lingual tasks, and scaling laws~\citep{zhu2023extrapolating}, or leverage self-distillation from high-resource languages~\citep{zhang2024enhancing}. Beyond data-level interventions, \citet{ye2025cctuning} explored latent-level cross-lingual interactions via a cross-lingual connection mechanism.
Empirical results consistently show that incorporating multilingual examples during instruction tuning leads to significantly stronger multilingual generalization compared to purely monolingual or English-centric approaches~\citep{shaham2024multilingual, chen-etal-2024-monolingual}.

\paragraph{Data Selection.}
A central goal of data selection in model fine-tuning is to minimize the amount of training data while maintaining—or even improving—model performance. Early work primarily focused on ensuring data quality and diversity. For example, \citet{touvron2023llama2} and \citet{zhou2023lima} employed elaborate filtering pipelines to emphasize the role of high-quality supervision. Data diversity has also gained increasing attention, as it is crucial for training robust and generalizable models~\citep{bukharin-etal-2024-data}. \citet{lu2023instag} proposed measuring diversity via intention tags of instructions, while \citet{yang2025diversity} achieved diversity-driven data selection through sparse autoencoder. Other selection criteria include features such as sequence length~\citep{zhao2024long}, task complexity~\citep{xu2024wizardlm}, etc.
Beyond data-intrinsic properties, a growing line of work adopts model-aware selection, identifying samples that align closely with the target domain based on gradient similarity or other interaction signals between training and target sets~\citep{xia2024less, zhao2025beyond}.

While these approaches have advanced instruction tuning through data-level and model-level interventions, little attention has been paid to the intrinsic linguistic structure present in multilingual training data. Our work addresses this gap by investigating language separability as a guiding signal for multilingual data selection.

\section{Measurement of Language Separability}

To quantify how well the training samples from different languages are separated in representation space, we use the silhouette score~\citep{rousseeuw1987silhouettes}. Originally designed to assess clustering quality, this metric favors compact intra-group and distant inter-group structures, making it well-suited for evaluating language separability.

Let $D = \{d_1, d_2, \dots, d_L\}$ denote a multilingual supervised instruction dataset, where each $d_l=\{p_i^l\}= \{(x_i^l, y_i^l)\}$ represents a language-specific subset corresponding to language $l \in \{1, \dots, L\}$.
Here, each sample pair $p_i^l$ consists of an input instruction $x_i^l$ and its corresponding ground-truth response $y_i^l$.
Given a model, we obtain the representation for each data point by formatting the instruction–response pair $(x_i^l, y_i^l)$ using the model's training template, feeding it into the model, and extracting the hidden state of the last token.
All representations are then grouped by language labels.

For each data point $p_i^l$, we compute its average distance $a(p_i^l)$ to all other points in the same language cluster $d_l$:
\begin{equation}
a(p_i^l) = \frac{1}{|d_l|-1}\sum_{p_j^l \in d_l, j \neq i}{\mathrm{dist}(p_i^l, p_j^l)}
\end{equation}
where $|d_l|$ denotes the number of data points in $d_l$, and $\mathrm{dist}(p_i^l, p_j^l)$ represents the Euclidean distance between $p_i^l$ and $p_j^l$ in the their representation space. The value of $a(p_i^l)$ quantifies how well the sample $p_i^l$ fits within its language cluster $d_l$ (the smaller the value, the better the alignment).

Next, we measure the dissimilarity between $p_i^l$ and other clusters by computing its average distance to the samples in its nearest neighboring cluster:
\begin{equation}
b(p_i^l) = \mathop{\operatorname{min}}_{m \neq l}\frac{1}{|d_m|}\sum_{p_j^m \in d_m}{\mathrm{dist}(p_i^l, p_j^m)}
\end{equation}
A larger $b(p_i^l)$ indicates that $p_i^l$ is more distinct from samples in other language clusters.

The silhouette score for each data point $p_i^l$ jointly considers intra-cluster compactness $a(p_i^l)$ and inter-cluster separation $b(p_i^l)$, and is defined as:
\begin{equation}
s(p_i^l) = \frac{b(p_i^l) - a(p_i^l)}{\max\{a(p_i^l),b(p_i^l)\}}
\end{equation}
The silhouette score ranges from $-1$ to $1$, with higher values indicating that the sample $p_i^l$ is well-clustered and clearly separated from other languages in the representation space.
Our language separability guided data pre-selection strategy prioritizes the top $\rho\%$ of samples with the highest silhouette scores within each language cluster, encouraging the model to form and maintain clearer linguistic boundaries by training on selected highly separable data.

\section{Experiments}\label{sec:exp}

\subsection{Setups}

\paragraph{Models.}
We selected two representative LLMs for our main experiments: (1) \textit{LLaMA-3.1-8B}~\citep{dubey2024llama} and (2) \textit{Qwen2.5-7B}~\citep{qwen2.5}.

\paragraph{Training Corpus.}
We totally select 97,696 multilingual instruction pairs from \textit{aya dataset}~\citep{singh2024aya} as our full training corpus and the training corpus covers 31 languages, ensuring extensive multilingual coverage (see detailed statistics in Appendix~\ref{app:training_data}). Our training processes are conducted on \textit{8 * A100-80GB} GPUs with the following settings: \textit{batch size=16}, \textit{epochs=3}, \textit{learning rate=1.0e-5}, \textit{warmup ratio=0.1}, and \textit{bf16=true}. The implementation is based on \textit{LLaMA-Factory}~\citep{zheng2024llamafactory}.

\paragraph{Evaluation Datasets.}
We conduct experiments on 6 benchmarks, which can be categorized into: 
\begin{itemize}[leftmargin=*]
\setlength{\parsep}{0pt}
\setlength{\parskip}{0pt}
\item \textbf{Multilingual Understanding:} (1) \textit{XNLI}~\citep{conneau2018xnli}, a multilingual natural language inference (NLI) dataset, (2) \textit{XStoryCloze}~\citep{lin-etal-2022-shot}, a multilingual commonsense reasoning dataset for evaluating story understanding and (3) \textit{MMMLU}, the multilingual version of \textit{MMLU}~\citep{hendrycks2020measuring}, designed to evaluate models' general knowledge.
\item \textbf{Multilingual Generation:} (1) \textit{MKQA}~\citep{longpre-etal-2021-mkqa}, an open-domain multilingual question answering evaluation dataset, (2) \textit{XQuAD}~\citep{artetxe-etal-2020-cross}, a question answering dataset and (3) \textit{XLSum}~\citep{hasan-etal-2021-xl}, a multilingual abstractive summarization benchmark comprising professionally annotated article-summary pairs.
\end{itemize}

\input{tables/main}

For each dataset, we evaluate on 10 languages, covering a total of 22 languages.
\textit{Accuracy} metric is used for \textit{XNLI}, \textit{XStoryCloze}, \textit{MMMLU}, \textit{MKQA} and \textit{XQuAD} datasets. And for \textit{XLSum} dataset, \textit{ROUGE-L} scores are reported.
We use greedy decoding with a max of 40 new tokens for each model. More details are provided in Appendix~\ref{app:eval_data},\ref{app:prompts}.

\paragraph{Baselines.} Detailed implementations are in Appendix~\ref{app:baselines}.
\begin{itemize}[leftmargin=*]
\setlength{\parsep}{0pt}
\setlength{\parskip}{0pt}
\item \textbf{Random \textit{(Rand)}}: samples are randomly selected from the full training set. Results are averaged over three different random seeds to ensure robustness.
\item \textbf{KMeans Clustering \textit{(KMC)}} performs kmeans in the embedding space of model \texttt{all-MiniLM-L6-v2} and selects the closest sample to each cluster centroid.
\item \textbf{Lexical Diversity \textit{(MTLD)}}~\citep{mccarthy2010mtld}: selects samples with higher lexical diversity.
\item \textbf{Naturalness \textit{(Nat)}}~\citep{zhong-etal-2022-towards} selects samples that better resemble natural human-written text.
\item \textbf{Coherence \textit{(Coh)}}~\citep{zhong-etal-2022-towards} selects samples where the response serves as a coherent continuation of the paired question.
\item \textbf{Understandability \textit{(Und)}}~\citep{zhong-etal-2022-towards} selects samples that are more understandable.
\item \textbf{Importance Resampling \textit{(DSIR)}}~\citep{xie2023data} selects samples most relevant to the target set by estimating their importance weights.
\item \textbf{Gradient Similarity \textit{(LESS)}}~\citep{xia2024less}: selects samples most relevant to the target set by low-rank gradient similarity search.
\end{itemize}
The target set for the target-dependent baselines \textbf{\textit{DSIR}} and \textbf{\textit{LESS}} is constructed by sampling 3 examples from each language subset across all 6 benchmarks.

\paragraph{Detailed Implementations of {\name}.}
As a pre-selection strategy, {\name} first selects the top $\rho\%$ of samples with the highest language separability scores from each language cluster in the original multilingual training set (with $\rho=20$ in our main experiments). These selected samples are then passed to existing data selection methods for further fine-grained filtering.
In our main experiments, we apply {\name} on top of both \textit{Random} selection and the three best-performing baselines for each model. We consider data selection settings of 1\%, 3\%, and 5\%—i.e., the final training set contains 1\%, 3\%, or 5\% of the full data—and report the average performance across these three settings.

\input{pictures/lang_gains}

\subsection{Main results}\label{sec:results}
Table~\ref{tab:main} reports average results across languages for each dataset, and Figure~\ref{fig:lang_gains} shows the average relative gains or declines on high- and low-resource languages.
The detailed results for each language can be found in Appendix~\ref{app:detailed_langs}.

\paragraph{(1) Existing Methods Are Not Universally Effective.} Existing feature-based and target-dependent data selection methods fail to deliver consistent improvements in joint multilingual instruction tuning.
For example, the trending method \textit{LESS} achieves the best performance on \textit{LLaMA-3.1-8B} ($\Delta=+4.90\%$), yet ranks the worst on \textit{Qwen2.5-7B} ($\Delta=-7.35\%$).
Moreover, \textit{Nat}, \textit{Coh}, and \textit{Und}—the three text quality-based selection methods—perform significantly worse on understanding tasks compared to generation tasks.
And on \textit{Qwen2.5-7B}, we observe that all baselines underperform even the \textit{Random} selection strategy.
These observations highlight the challenge of designing universally effective selection strategies in multilingual settings.

\paragraph{(2) {\name} Yields a Performance Boost, Especially on Understanding Tasks and Low-Resource Languages.}
Overall, applying {\name} on top of existing selection methods achieves consistent performance improvements, as evidenced by the increase in $\Delta$ values. And the gains are more pronounced on understanding tasks.
Besides, we also observe that for \textit{LLaMA-3.1-8B}, which exhibits weaker inherent multilingual capabilities, both the baselines and {\name} yield more substantial improvements compared to \textit{Qwen2.5-7B}, whose multilingual competence is already stronger.
In Figure~\ref{fig:lang_gains}, we present the average performance gains or declines when applying {\name}, separately for high-resource and low-resource languages\protect\footnotemark.
Overall, {\name} brings positive improvements on both language groups and the gains are more pronounced on low-resource languages, indicating that {\name} can help mitigate the performance gap between high- and low-resource languages.

\footnotetext{We follow the taxonomy in \url{https://microsoft.github.io/linguisticdiversity/assets/lang2tax.txt}, where languages rated 4 or 5 are considered relatively high-resource, and others low-resource. And since the 10 languages selected from \textit{MKQA} are not typical low-resource, \textit{MKQA} dataset is excluded from the figure.}

\section{Further Analysis}

\subsection{Analysis of the Pre-selection Ratio $\rho$}

We vary the pre-selection ratio $\rho$ from 10\% to 100\% and report the performance on \textit{MMMLU}, \textit{XLSum} in Figure~\ref{fig:pre_percent} (5\% setting, \textit{LLaMA-3.1-8B}) using both \textit{Rand} and \textit{LESS} as downstream selectors. Note that when $\rho=100\%$, {\name}\textit{+baseline} becomes equivalent to the vanilla baseline.

We observe that {\name} is not highly sensitive to the choice of $\rho$, and generally performs well when $\rho$ is in the range of 20\% to 70\%.
A very small $\rho$ (e.g., 10\%) tends to over-focus on highly separable samples, leading to reduced diversity (as shown in Figure~\ref{fig:pilot} (b)), and potential performance drops. On the other hand, as $\rho$ approaches 100\%, the pre-selection effect of {\name} gradually converges toward those of the vanilla baselines.
Moreover, the choice of $\rho$ also presents a trade-off between data diversity and selection efficiency. A smaller $\rho$ means that subsequent selection methods—especially those with higher computational cost (e.g., \textit{LESS})—operate on a much smaller candidate pool, reducing overhead at the cost of potentially reduced diversity.

\input{pictures/pre_percent}

\paragraph{Note that computational cost is analyzed in Appendix~\ref{app:cost}.}

\subsection{Analysis of Linguistic Boundary Representation}
To provide a more intuitive understanding of {\name}, we employ t-SNE~\citep{van2008visualizing} to visualize the representations of 200 sentences sampled from \textit{XNLI} in parallel across English, Chinese and Arabic.

\begin{figure*}[t]
    \centering
    \includegraphics[width=1\textwidth]{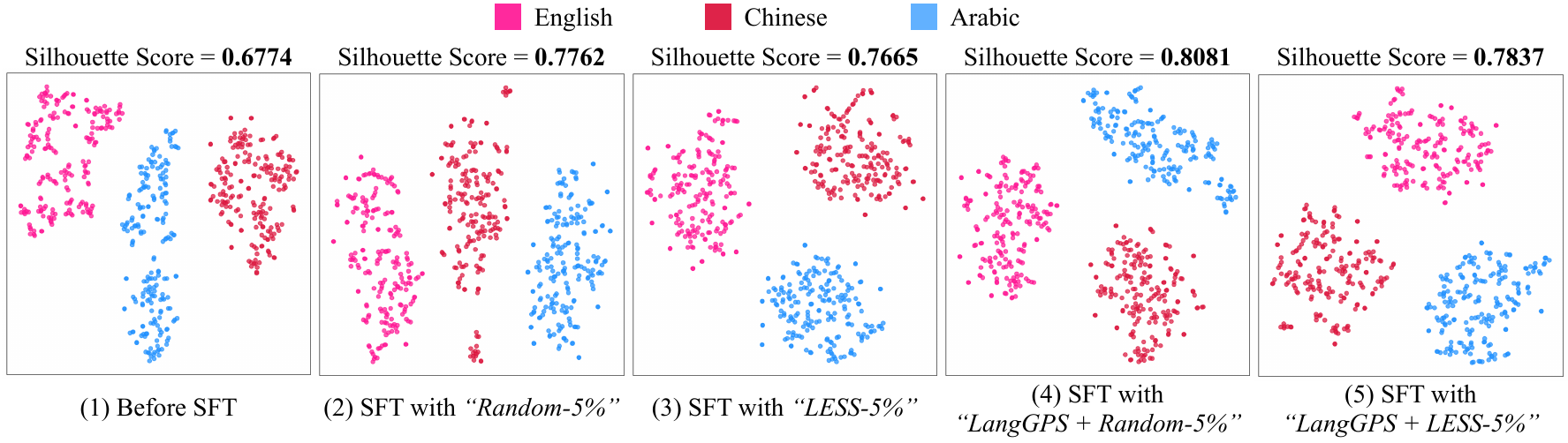}
    \caption{t-SNE visualizations of output representations from \textit{LLaMA-3.1-8B} under different training settings: before SFT, after SFT with \textit{Rand-5\%}, \textit{LESS-5\%}, \textit{LangGPS+Rand-5\%}, and \textit{LangGPS+LESS-5\%}. Average silhouette scores are also reported.
    }
    \label{fig:tSNE}
\end{figure*}

As shown in Figure~\ref{fig:tSNE} (1) (2) (3), multilingual SFT leads to clearer linguistic boundaries and more compact clusters in the model’s representation space, reflected by increased average silhouette scores. Moreover, comparing Figure~\ref{fig:tSNE} (2) vs. (4) and (3) vs. (5), we observe that applying {\name} further amplifies this effect: models trained with highly separable samples exhibit even higher silhouette scores, suggesting that {\name} helps the model better form and maintain clear language boundaries.

\input{pictures/translation}

\subsection{The Role of Low-Separability Samples}\label{sec:low_sep}
Our experiments show that prioritizing highly separable samples during training improves multilingual performance. But \textit{\textbf{are low-separability samples entirely unhelpful?}}

Low-separability samples often entangle with other language clusters in representation space, lacking clear language-specific features—but they may instead  serve as implicit bridges for cross-lingual alignment and information sharing.
To investigate this, we revisit the setup in Figure~\ref{fig:pilot} (a) and Appendix~\ref{app:pre_exp}, training models with (1) top separability-scoring, (2) bottom separability-scoring, and (3) randomly selected samples (with varying sizes), and evaluate them on the \textit{FLORES+} translation task to assess the role of low-separability data in cross-lingual scenarios.

The results in Figure~\ref{fig:translation}, together with Figure~\ref{fig:pilot}(a), reveal several observations in common:
\textbf{(1)} The three settings differ notably when training with limited samples, but their performances converge as training data increases.
\textbf{(2)} Low-separability samples lead to a severe cold-start problem—when trained on just 50 examples, models perform significantly worse than those trained on high-separability or random data, likely due to the lack of clear language-specific signals in a small set of entangled samples.

Furthermore, we observe that the trends differ significantly between the ``$X \rightarrow En$'' and ``$En \rightarrow X$'' translation directions.
\textbf{(1)} In ``$X \rightarrow En$'', none of the training settings yields improvement in the stable phase (red shaded area)—suggesting that vanilla multilingual SFT offers limited benefit for $X \rightarrow En$ translation. Nevertheless, low-separability samples result in the least degradation and occasionally yield marginal gains (e.g., at 400, 500, or 3000 samples).  
\textbf{(2)} In contrast, the overall trend of ``En $\rightarrow$ X'' direction shows clear gains from multilingual SFT.
And within the 200-2000 sample dip (the cyan-shaded area), model trained with low-separability samples experiences a milder decline compared to those trained with high-separability samples.

\paragraph{Summary.} With limited training data, low-separability samples—due to their entangled representations across languages—may contribute more to cross-lingual alignment rather than supporting language-specific modeling, whereas high-separability samples are more effective for learning language-specific features and enabling a better warm start.

\input{tables/curriculum}

\subsection{Language Separability Guided Multilingual Curriculum Learning}
Curriculum learning shares an underlying intuition with data selection: both aim to improve model training by leveraging differences in sample utility.
While data selection focuses on identifying and retaining a subset of training samples deemed beneficial, curriculum learning instead retains the full dataset but modulates the order in which the samples are presented. In essence, both strategies prioritize how the model interacts with training data—either by selecting what to learn from, or by deciding when to learn from it.

In this section, we explore whether language separability can serve as a guiding signal for multilingual curriculum learning. We design three curriculum strategies based on language separability:
(1) \textit{Ascending}: training progresses from low to high separability samples;
(2) \textit{Descending}: from high to low separability;
(3) \textit{Balanced}: training data are interleaved across different separability ranges to maintain an approximately uniform separability distribution throughout training (detailed implementations and algorithm are presented in Appendix~\ref{app:curriculum}).

The results in Table~\ref{tab:curriculum} show that naively presenting data in strictly ascending or descending separability order leads to unsatisfactory performance. Instead, the \textit{Balanced} strategy, where samples of varying separability are mixed evenly throughout training, yields the most stable and generalizable improvements across different models and datasets.
This finding suggests that maintaining a diversity of separability levels during training helps the model benefit from both highly structured (language-specific) and entangled (cross-lingual) samples, striking a better trade-off.

\section{Conclusion}

We present {\name}, a lightweight and broadly compatible data pre-selection framework that leverages language separability as the guiding signal for multilingual data selection.
By quantifying how well different languages are distinguished in the model’s representation space, {\name} prioritizes highly separable samples that help models form clearer linguistic boundaries.
Extensive experiments show that applying {\name} on top of existing selection methods improves their effectiveness and generalizability in multilingual training, especially for understanding tasks and low-resource languages.
Further analysis highlights the complementary roles of high- and low-separability samples, and demonstrates that language separability also benefits multilingual curriculum learning.
Our findings position language separability as a principled signal for organizing multilingual data, offering a new perspective for improving the effectiveness and generalization of multilingual LLMs.

\section*{Acknowledgements}
Xiaocheng Feng, Xiachong Feng and Dandan Tu are the co-corresponding authors of this work.
We thank the anonymous reviewers for their insightful comments.
This work was supported by the National Natural Science Foundation of China (NSFC) (grant 62522603, 62276078, U22B2059), the Key R\&D Program of Heilongjiang via grant 2022ZX01A32, and the Fundamental Research Funds for the Central Universities ( XNJKKGYDJ2024013 ).

\paragraph{Limitations of this work are discussed in Appendix~\ref{app:limit}.}

\bibliography{aaai2026}


\appendix

\section{Experiment Details}

\subsection{Preliminary Experiments}\label{app:pre_exp}

We conduct preliminary experiments by fine-tuning \textit{LLaMA-3.1-8B} using three types of training samples: (1) \textbf{High-separability} samples with the highest separability scores, (2) \textbf{Low-separability} samples with the lowest separability scores, and (3) \textbf{Random} selected samples from the full dataset. \textbf{To ensure fair comparison}, we select samples in accordance with the original language distribution of the dataset. For the random setting, we perform three independent samplings and report the averaged results. We vary the training data size from 0 to 5,000 to observe the impact of separability under different data scales.
The full training corpus and \textit{MMMLU} dataset used in preliminary experiments follows the settings in Appendix~\ref{app:training_data},\ref{app:eval_data}.

Besides, we also present t-SNE visualizations of samples selected by different strategies under the same language distribution, using the last-token hidden states from \textit{LLaMA-3.1-8B} as representations.

\subsection{Training Corpus}\label{app:training_data}

We totally select 97,696 multilingual instruction pairs from \textit{aya dataset}~\citep{singh2024aya} as our full training corpus and the training corpus covers 31 languages, ensuring extensive multilingual coverage. Detailed statistics are as follows:
\begin{lstlisting}[language=Python]
{
  "Basque": 939,
  "Bengali": 1534,
  "Egyptian Arabic": 529,
  "English": 3944,
  "French": 1422,
  "German": 241,
  "Greek": 623,
  "Haitian": 106,
  "Hindi": 1153,
  "Indonesian": 786,
  "Italian": 738,
  "Japanese": 6259,
  "Korean": 361,
  "Moroccan Arabic": 8090,
  "Najdi Arabic": 136,
  "Portuguese": 8997,
  "Russian": 423,
  "Simplified Chinese": 3038,
  "South Levantine Arabic": 81,
  "Spanish": 3854,
  "Standard Arabic": 4995,
  "Swahili": 366,
  "Ta'izzi-Adeni Arabic": 129,
  "Tamil": 14133,
  "Telugu": 8439,
  "Thai": 724,
  "Turkish": 4046,
  "Ukrainian": 522,
  "Urdu": 654,
  "Vietnamese": 8676,
  "Yoruba": 11758
}
\end{lstlisting}

\subsection{Evaluation Datasets}\label{app:eval_data}

We conduct experiments on 6 benchmarks, which can be categorized into: 
\begin{itemize}[leftmargin=*]
\setlength{\parsep}{0pt}
\setlength{\parskip}{0pt}
\item \textbf{Multilingual Understanding:} (1) \textit{XNLI}~\citep{conneau2018xnli}, a multilingual natural language inference (NLI) dataset, (2) \textit{XStoryCloze}~\citep{lin-etal-2022-shot}, a multilingual commonsense reasoning dataset for evaluating story understanding and (3) \textit{MMMLU}, the multilingual version of \textit{MMLU}~\citep{hendrycks2020measuring}, designed to evaluate models' general knowledge.
\item \textbf{Multilingual Generation:} (1) \textit{MKQA}~\citep{longpre-etal-2021-mkqa}, an open-domain multilingual question answering evaluation dataset, (2) \textit{XQuAD}~\citep{artetxe-etal-2020-cross}, a question answering dataset and (3) \textit{XLSum}~\citep{hasan-etal-2021-xl}, a multilingual abstractive summarization benchmark comprising professionally annotated article-summary pairs.
\end{itemize}

For each dataset, we conduct experiments on 10 language, covering a total of 22 languages.
For \textit{XNLI}, \textit{XStoryCloze}, \textit{MMMLU}, \textit{MKQA} and \textit{XQuAD} datasets, \textit{Accuracy} metric is used for evaluation. And for \textit{XLSum} dataset, \textit{ROUGE-L} scores are reported.
We use greedy decoding with a max of 40 new tokens for each model.

\textit{XNLI}, \textit{XStoryCloze}, and \textit{MMMLU} all belong to the multiple-choice category. For these datasets, a model's response is considered correct only if it contains the correct option and excludes all other options.
For the short QA generative dataset \textit{MKQA} and \textit{XQuAD}, a model's answer is deemed correct if the gold answer appears in the model's response.
\begin{promptbox}

    \textbf{Involved Languages (10 languages each dataset)}
    \\
    \textit{XNLI:} en, ar, el, es, fr, hi, sw, tr, vi, zh\\
    \textit{XStoryCloze:} en, ar, es, eu, hi, id, ru, sw, te, zh \\
    \textit{MMMLU:} en, ar, bn, de, es, fr, hi, sw, yo, zh \\
    \textit{MKQA:} en, ar, de, ja, ko, pt, ru, tr, vi, zh \\
    \textit{XQuAD:} en, ar, de, el, es, hi, ru, th, tr, zh \\
    \textit{XLSum:} en, es, fr, ko, pt, sw, tr, uk, vi, zh
\end{promptbox}
\begin{promptbox}

    \textbf{Sample Size}
    \\
    \text{\textit{XNLI:} $1000 \times 10 = 10000$} (parallel)\\
    \text{\textit{XStoryCloze:} $1511 \times 10 = 15110$} (parallel)\\
    \text{\textit{MMMLU:} $1000 \times 10 = 10000$} (parallel)\\
    \text{\textit{MKQA:} $1000 \times 10 = 10000$} (parallel)\\
    \text{\textit{XQuAD:} $1190 \times 10 = 11900$} (parallel)\\
    \text{\textit{XLSum:} $100 \times 10 = 1000$} (non-parallel)
\end{promptbox}

\subsection{Prompts}\label{app:prompts}

\input{tables/time}

\input{tables/prompts}
The prompts we used for each datasets are listed in Table~\ref{tab:prompts}.

\subsection{Baselines Settings}\label{app:baselines}

The implementations of the baseline methods involved.

\paragraph{Baselines.}
\begin{itemize}[leftmargin=*]
\setlength{\parsep}{0pt}
\setlength{\parskip}{0pt}
\item \textbf{Random \textit{(Rand)}}: samples are randomly selected from the full training set. Results are averaged over three different random seeds to ensure robustness.
\item \textbf{KMeans Clustering \textit{(KMC)}} performs k-means in the embedding space of model \texttt{all-MiniLM-L6-v2}\footnote{We use the English encoder \texttt{all-MiniLM-L6-v2} instead of a multilingual encoder because multilingual encoders are often trained to align semantically similar sentences across languages into a shared embedding space. While this design benefits cross-lingual retrieval or transfer learning tasks, it also has a notable side effect: they tend to suppress language-specific information. That is, samples from different languages but with similar semantics may be embedded closely together, making it difficult to distinguish languages based solely on their representations.
In our context—where language-specific structure and diversity are important signals for data selection—such alignment behavior may be counterproductive. In contrast, monolingual models like \texttt{all-MiniLM-L6-v2} retain more surface-level and syntactic features, making them more suitable for preserving instructional variation and language-specific patterns.} and selects the closest sample to each cluster centroid. \texttt{all-MiniLM-L6-v2} maps sentences \& paragraphs to a 384 dimensional dense vector space. To select $k$ representative samples, we perform 
k-means clustering in this embedding space and set the number of clusters in k-means to $k$. And then we identify the sample closest to each cluster centroid as the selected instance.
\item \textbf{Lexical Diversity \textit{(MTLD)}}~\citep{mccarthy2010mtld}: selects samples with higher lexical diversity. We use LTP~\citep{che2020n} for word segmentation and the \texttt{lexical-diversity} Python package for computing MTLD scores, and select the top-scoring samples for training.
\item \textbf{Naturalness \textit{(Nat)}}~\citep{zhong-etal-2022-towards} selects samples that better resemble natural human-written text. We utilize \texttt{unieval-dialog}~\citep{zhong-etal-2022-towards} for computing naturalness scores, and select the top-scoring samples for training.
\item \textbf{Coherence \textit{(Coh)}}~\citep{zhong-etal-2022-towards} selects samples where the response serves as a coherent continuation of the paired question. We utilize \texttt{unieval-dialog}~\citep{zhong-etal-2022-towards} for computing coherence scores, and select the top-scoring samples for training.
\item \textbf{Understandability \textit{(Und)}}~\citep{zhong-etal-2022-towards} selects samples that are more understandable.
We utilize \texttt{unieval-dialog}~\citep{zhong-etal-2022-towards} for computing understandability scores, and select the top-scoring samples for training.
\item \textbf{Importance Resampling \textit{(DSIR)}}~\citep{xie2023data} selects samples most relevant to the target set by estimating their importance weights.
\item \textbf{Gradient Similarity \textit{(LESS)}}~\citep{xia2024less}: selects samples most relevant to the target set by low-rank gradient similarity search.
\end{itemize}
The target set for the target-dependent baselines \textbf{\textit{DSIR}} and \textbf{\textit{LESS}} is constructed by sampling 3 examples from each language subset across all 6 benchmarks.

\subsection{Detailed Experimental Results on Each Involved Language}\label{app:detailed_langs}

In this section, we present the detailed results for each language involved in Table~\ref{tab:main_xnli}, \ref{tab:main_xstorycloze}, \ref{tab:main_mmmlu}, \ref{tab:main_mkqa}, \ref{tab:main_xquad}, \ref{tab:main_xlsum}.

\section{Computational Cost Analysis}\label{app:cost}
In this section, we report the actual time cost of {\name} and other baselines in Table~\ref{tab:cost}, measured as \textbf{single} A100 GPU hours on full training corpus.

We observe that the majority of {\name}'s computational overhead comes from the \textit{Getting Representation} step, where we pass the training samples into the model and get the representations of its last input token, while the time spent on \textit{Silhouette Scoring} and \textit{Data Selection} is negligible in practice.
Notably, though \textit{Getting Representation} costs hours, the process itself is quite simple and implementation-friendly—requiring only a single forward pass\footnote{\texttt{model(**inputs, output\_hidden\_states=True)}} over the training corpus without any additional computation.

Besides, as a pre-selection strategy, {\name} can help reduce the computational overhead of subsequent selection methods—particularly those with higher costs, such as \textit{LESS}—by narrowing down the candidate pool. Specifically, under a pre-selection ratio of $\rho=20\%$, the total cost of {\name}\textit{+LESS} becomes:
$$
\begin{aligned}
    &\rm{Cost}({\name}\textit{+LESS}) = \rm{Cost}({\name}_{total}) \\
    & \quad + \rm{Cost}(\textit{LESS}_{Warmup})  \\
    & \quad + {\textbf{20\%}} \times \rm{Cost}(\textit{LESS}_{Gradient}) \\
    & \quad \approx 5.2 \ \rm{Hours} < Cost(\textit{LESS}_{total}) = 8 \ \rm{Hours}
\end{aligned}
$$

\section{Curriculum Learning Implementation}\label{app:curriculum}

We first sort the entire training corpus by language separability scores and divide it into 10 equally sized buckets based on score percentiles: Top 0–10\%, Top 10–20\%, ..., Top 90–100\%. Specifically, each bucket contains the samples within the corresponding range of separability scores from each language cluster in the original multilingual training set. Then, we apply the following curriculum learning strategies:

\begin{itemize}[leftmargin=*]
\setlength{\parsep}{0pt}
\setlength{\parskip}{0pt}
\item \textbf{Ascending:} Train the model starting from the samples with the \emph{lowest} separability scores and gradually proceed to those with the \emph{highest}. (Training order: Top 90–100\%, 80–90\%, \ldots, 0–10\%)
\item \textbf{Descending:} Train the model starting from the samples with the \emph{highest} separability scores and gradually proceed to those with the \emph{lowest}. (Training order: Top 0–10\%, 10–20\%, \ldots, 90–100\%)
\item \textbf{Balanced:} Training data are interleaved across different separability ranges to maintain an approximately uniform separability distribution throughout training, ensuring no score range is over- or under-represented. The detailed algorithm is as follows:
\end{itemize}

\begin{algorithm}[H]
\caption{Balanced Training Data Construction}
\begin{algorithmic}[1]
\State Initialize training set $\mathcal{D}_{\text{train}} \gets [\;]$
\State Sort all training samples by language separability scores.
\State Divide them into 10 equal-sized buckets: $\mathcal{B}_1$ (Top 0-10\%), $\mathcal{B}_2$ (Top 10-20\%), $\dots$, $\mathcal{B}_{10}$ (Top 90-100\%), where each $\mathcal{B}_i$ contains the corresponding score-range samples from each language cluster.
\While{any bucket $\mathcal{B}_i$ is non-empty}
    \State Initialize mini-batch $\mathcal{M} \gets [\;]$
    \For{$i = 1$ to $10$}
        \If{$\mathcal{B}_i$ is not empty}
            \State Sample one instance $x$ from $\mathcal{B}_i$
            \State Append $x$ to $\mathcal{M}$
        \EndIf
    \EndFor
    \State Shuffle $\mathcal{M}$ and append to $\mathcal{D}_{\text{train}}$
\EndWhile
\State \Return $\mathcal{D}_{\text{train}}$
\end{algorithmic}
\end{algorithm}

\section{Limitation}\label{app:limit}
This work exhibits several limitations worth noting.
Firstly, the quantification of language separability in {\name} essentially reflects the model’s own multilingual modeling capability. As multilingual capacities vary across models, {\name} requires model-specific data selection, lacking the simplicity of a one-size-fits-all solution.
Secondly, our experiments were conducted on \textit{LLaMA-3.1-8B} and \textit{Qwen2.5-7B}. While these models represent important milestones in open-source LLM development, the evaluation across more LLMs would improve the generalizability of our findings across the broader LLM ecosystem.
Thirdly, in Section~\ref{sec:low_sep}, we examine the impact of high- and low-separability samples on translation performance. It is worth noting that the training data (aya dataset) is not designed to improve translation quality. Hence, the goal of this section is not to show that low-separability samples enhance translation performance. Rather, we highlight that the milder decline of low-separability samples within a fluctuating performance region reveals their distinct functional role compared to high-separability samples.

\input{tables/main_xnli}
\input{tables/main_xstorycloze}
\input{tables/main_mmmlu}
\input{tables/main_mkqa}
\input{tables/main_xquad}
\input{tables/main_xlsum}

\end{document}

%% file: tables/main.tex
\begin{table*}[p]
\centering
\setlength{\aboverulesep}{1pt} 
\setlength{\belowrulesep}{1pt} 
\renewcommand{\arraystretch}{1.1} 
\setlength{\tabcolsep}{1.5pt}

\resizebox{\linewidth}{!}{
    \begin{tabular}{lcccc|cccc|cccc|cccc|cccc|cccc|r}
    \toprule
    \multirow{3}[3]{*}{\textbf{Method}} & \multicolumn{24}{c}{\textbf{\textit{Model: LLaMA-3.1-8B}}} \\
    \cmidrule(lr){2-25}
     & \multicolumn{12}{c}{\textbf{Multilingual Understanding}} & \multicolumn{12}{c}{\textbf{Multilingual Generation}} \\
    \cmidrule(lr){2-13}
    \cmidrule(lr){14-25}
    & \multicolumn{4}{c}{\textbf{XNLI}} & \multicolumn{4}{c}{\textbf{XStoryCloze}} & \multicolumn{4}{c}{\textbf{MMMLU}} & \multicolumn{4}{c}{\textbf{MKQA}} & \multicolumn{4}{c}{\textbf{XQuAD}} & \multicolumn{4}{c}{\textbf{XLSum}} \\
    \cmidrule(lr){1-25}
    \textit{Full} & \multicolumn{4}{c}{36.6} & \multicolumn{4}{c}{64.2} & \multicolumn{4}{c}{37.7} & \multicolumn{4}{c}{15.1} & \multicolumn{4}{c}{60.8} & \multicolumn{4}{c}{23.3} \\
    \midrule
    (\%) & {1\%} & {3\%} & {5\%} & {\textbf{Avg.}} & {1\%} & {3\%} & {5\%} & {\textbf{Avg.}} &  {1\%} & {3\%} & {5\%} & {\textbf{Avg.}} & {1\%} & {3\%} & {5\%} & {\textbf{Avg.}} & {1\%} & {3\%} & {5\%} & {\textbf{Avg.}} &  {1\%} & {3\%} & {5\%} & {\textbf{Avg.}} & \multicolumn{1}{c}{$\Delta$} \\
    \midrule
    \textit{Rand} & 25.1 & 33.0 & 35.4 & 31.2 & 43.7 & 61.4 & 61.1 & 55.4 & 34.2 & 37.4 & 38.1 & 36.5 & 14.6 & 14.4 & 14.4 & 14.5 & 63.6 & 59.9 & 59.6 & 61.1 & 21.9 & 22.8 & 22.8 & 22.5 & $-6.44\%$ \\
    \midrule
    \multicolumn{25}{c}{\textbf{\textit{Feature-based Baselines}}} \\
    
    \textit{KMC*} & 24.3 & 35.5 & 34.1 & 31.3 & 44.1 & 63.7 & 65.6 & 57.8 & 38.1 & 39.6 & 37.2 & 38.3 & 15.1 & 14.8 & 15.0 & 15.0 & 59.7 & 58.9 & 59.3 & 59.3 & 21.0 & 23.1 & 22.4 & 22.2 & $-5.18\%$ \\
    \textit{MTLD} & 28.8 & 34.9 & 34.4 & 32.7 & 34.3 & 52.9 & 47.1 & 44.8 & 25.1 & 32.5 & 29.9 & 29.1 & 15.7 & 15.9 & 14.5 & 15.4 & 60.3 & 62.5 & 57.2 & 60.0 & 19.5 & 21.9 & 24.2 & 21.8 & $-11.55\%$ \\
    \textit{Nat} & 20.4 & 40.1 & 36.2 & 32.2 & 13.3 & 40.0 & 54.8 & 36.0 & 24.5 & 35.3 & 37.6 & 32.4 & 18.5 & 17.3 & 17.1 & 17.6 & 59.3 & 61.3 & 59.5 & 60.0 & 19.6 & 20.9 & 22.1 & 20.8 & $-10.80\%$ \\
    \textit{Coh*} & 28.3 & 37.1 & 36.3 & 33.9 & 42.1 & 54.5 & 51.5 & 49.4 & 34.6 & 35.9 & 36.4 & 35.6 & 18.2 & 18.1 & 18.4 & {\underline{18.2}} & 64.1 & 64.6 & 64.8 & 64.5 & 22.4 & 24.0 & 23.8 & {\underline{23.4}} & $-1.47\%$\\
    \textit{Und} & 23.5 & 38.1 & 35.0 & 32.2 & 25.3 & 47.5 & 59.6 & 44.1 & 28.8 & 35.0 & 38.4 & 34.1 & 18.2 & 16.9 & 17.4 & 17.5 & 60.6 & 60.6 & 59.5 & 60.2 & 20.6 & 21.2 & 22.2 & 21.3 & $-7.70\%$ \\
    \midrule
    \multicolumn{25}{c}{\textbf{\textit{Target-dependent Baselines}}} \\
    \textit{DSIR} & 29.5 & 27.7 & 36.9 & 31.4 & 36.7 & 39.7 & 49.5 & 42.0 & 27.3 & 30.1 & 35.1 & 30.8 & 17.7 & 16.9 & 16.4 & 17.0 & 64.7 & 63.4 & 62.4 & 63.5 & 22.1 & 23.1 & 23.5 & 22.9 & $-8.66\%$\\
    \textit{LESS*} & 33.9 & 33.7 & 37.8 & {\underline{35.1}} & 65.5 & 71.2 & 68.3 & {\underline{68.3}} & 40.4 & 40.6 & 40.8 & {\underline{40.6}} & 18.2 & 18.0 & 17.2 & {17.8} & 65.2 & 65.6 & 64.8 & {\underline{65.2}} & 21.2 & 21.7 & 22.8 & 21.9 & $\underline{+4.90\%}$ \\
    \midrule
    \multicolumn{25}{c}{\textbf{\textit{Applying LangGPS (to random selection and three best well-performing baselines marked with *)}}} \\
    \textit{Rand} & 26.4 & 34.7 & 36.2 & 32.5 & 48.7 & 72.5 & 72.5 & 64.6 & 33.7 & 38.2 & 39.7 & 37.2 & 15.4 & 15.0 & 14.7 & 15.0 & 62.6 & 60.2 & 62.0 & 61.6 & 21.9 & 22.8 & 23.8 & 22.9 & $-2.21\%$ \\
    & & & & \cellcolor{cyan!15}{(+1.3)} & & & & \cellcolor{cyan!15}(+9.2) & & & & \cellcolor{cyan!15}(+0.8) & & & & \cellcolor{cyan!15}(+0.5) & & & & \cellcolor{cyan!15}(+0.5) & & & & \cellcolor{cyan!15}(+0.4) \\
    \textit{KMC} & 26.1 & 38.0 & 37.6 & 33.9 & 42.4 & 66.3 & 69.2 & 59.3 & 38.3 & 40.2 & 38.8 & 39.1 & 15.3 & 15.1 & 13.8 & 14.7 & 61.8 & 60.3 & 60.6 & 60.9 & 21.5 & 22.2 & 23.8 & 22.5 & $-2.86\%$ \\
    & & & & \cellcolor{cyan!15}(+2.6) & & & & \cellcolor{cyan!15}(+1.5) & & & & \cellcolor{cyan!15}(+0.7) & & & & \cellcolor{gray!15}{(-0.3)} & & & & \cellcolor{cyan!15}(+1.6) & & & & \cellcolor{cyan!15}(+0.3)\\
    \textit{Coh} & 33.7 & 32.8 & 38.1 & 34.9 & 47.1 & 47.2 & 60.8 & 51.7 & 37.6 & 35.4 & 37.5 & 36.9 & 19.1 & 18.2 & 17.6 & {\textbf{18.3}} & 63.4 & 62.6 & 63.7 & 63.2 & 22.2 & 24.3 & 24.3 & {\textbf{23.6}} & $-0.01\%$\\
    & & & & \cellcolor{cyan!15}(+1.0) & & & & \cellcolor{cyan!15}(+2.3) & & & & \cellcolor{cyan!15}(+1.3) & & & & \cellcolor{cyan!15}(+0.1) & & & & \cellcolor{gray!15}(-1.3) & & & & \cellcolor{cyan!15}(+0.2) \\
    \textit{LESS} & 35.6 & 37.8 & 37.7 & {\textbf{37.0}} & 66.1 & 76.8 & 74.0 & {\textbf{72.3}} & 39.9 & 40.0 & 42.2 & {\textbf{40.7}} &17.1 & 16.5 & 17.0 & 16.9 & 65.6 & 64.3 & 64.3 & {\textbf{64.7}} & 22.5 & 22.8 & 23.4 & 22.9 & {$\bm{+6.37\%}$} \\
    & & & & \cellcolor{cyan!15}(+1.9) & & & & \cellcolor{cyan!15}(+4.0) & & & & \cellcolor{cyan!15}(+0.1) & & & & \cellcolor{gray!15}(-0.9) & & & & \cellcolor{gray!15}(-0.5) & & & & \cellcolor{cyan!15}(+1.0) \\
    \bottomrule
    \end{tabular}
}
\resizebox{\linewidth}{!}{
    \begin{tabular}{lcccc|cccc|cccc|cccc|cccc|cccc|r}
    \toprule
    \multirow{3}[3]{*}{\textbf{Method}} & \multicolumn{24}{c}{\textbf{\textit{Model: Qwen2.5-7B}}} \\
    \cmidrule(lr){2-25}
     & \multicolumn{12}{c}{\textbf{Multilingual Understanding}} & \multicolumn{12}{c}{\textbf{Multilingual Generation}} \\
    \cmidrule(lr){2-13}
    \cmidrule(lr){14-25}
    & \multicolumn{4}{c}{\textbf{XNLI}} & \multicolumn{4}{c}{\textbf{XStoryCloze}} & \multicolumn{4}{c}{\textbf{MMMLU}} & \multicolumn{4}{c}{\textbf{MKQA}} & \multicolumn{4}{c}{\textbf{XQuAD}} & \multicolumn{4}{c}{\textbf{XLSum}} \\
    \cmidrule(lr){1-25}
    \textit{Full} & \multicolumn{4}{c}{50.8} & \multicolumn{4}{c}{64.1} & \multicolumn{4}{c}{43.2} & \multicolumn{4}{c}{14.3} & \multicolumn{4}{c}{62.0} & \multicolumn{4}{c}{21.8} \\
    \midrule
    (\%) & {1\%} & {3\%} & {5\%} & {\textbf{Avg.}} & {1\%} & {3\%} & {5\%} & {\textbf{Avg.}} &  {1\%} & {3\%} & {5\%} & {\textbf{Avg.}} & {1\%} & {3\%} & {5\%} & {\textbf{Avg.}} & {1\%} & {3\%} & {5\%} & {\textbf{Avg.}} &  {1\%} & {3\%} & {5\%} & {\textbf{Avg.}} & \multicolumn{1}{c}{$\Delta$} \\
    \midrule
    \textit{Rand} & 45.6 & 53.0 & 54.7 & 51.1 & 54.2 & 76.0 & 72.7 & 67.7 & 49.3 & 45.6 & 44.1 & {\underline{46.4}} & 15.5 & 15.1 & 14.5 & {\underline{15.0}} & 65.4 & 64.0 & 63.9 & 64.4 & 22.1 & 22.2 & 22.7 & {\underline{22.3}} & {$\underline{+4.18\%}$}\\
    \midrule
    \multicolumn{25}{c}{\textbf{\textit{Feature-based Baselines}}} \\
    
    \textit{KMC*} & 51.1 & 49.8 & 51.3 & 50.8 & 74.3 & 64.4 & 73.0 & {\underline{70.6}} & 43.9 & 43.6 & 46.8 & 44.8 & 13.1 & 14.2 & 13.4 & 13.6 & 61.2 & 62.0 & 62.4 & 61.9 & 22.5 & 22.0 & 21.8 & 21.9 & $+1.52\%$\\
    \textit{MTLD*} & 50.1 & 53.9 & 51.5 & {\underline{51.8}} & 64.2 & 73.6 & 72.0 & 69.9 & 46.5 & 46.1 & 43.7 & 45.4 & 15.2 & 14.2 & 14.6 & 14.7 & 65.5 & 63.5 & 62.8 & 63.9 & 21.7 & 22.6 & 22.4 & 22.2 & $+4.03\%$\\
    \textit{Nat} & 37.9 & 42.4 & 49.1 & 43.1 & 56.7 & 51.3 & 58.8 & 55.6 & 46.5 & 44.1 & 42.5 & 44.4 & 15.2 & 14.5 & 14.4 & 14.7 & 59.2 & 62.2 & 61.5 & 60.9 & 21.4 & 20.9 & 20.5 & 20.9 & $-4.72\%$ \\
    \textit{Coh} & 38.4 & 50.3 & 44.0 & 44.2 & 63.7 & 61.4 & 53.6 & 59.6 & 40.7 & 44.5 & 38.0 & 41.0 & 15.3 & 14.7 & 14.5 & 14.8 & 60.2 & 61.8 & 61.1 & 61.0 & 21.7 & 21.8 & 22.4 & 22.0 & $-3.66\%$\\
    \textit{Und} & 39.5 & 48.8 & 49.3 & 45.9 & 49.1 & 57.7 & 58.5 & 55.1 & 43.8 & 45.1 & 42.2 & 43.7 & 14.9 & 14.4 & 14.7 & 14.7 & 58.3 & 61.8 & 62.2 & 60.8 & 21.1 & 20.3 & 20.9 & 20.8 & $-4.44\%$ \\
    \midrule
    \multicolumn{25}{c}{\textbf{\textit{Target-dependent Baselines}}} \\
    \textit{DSIR*} & 53.5 & 51.4 & 47.2 & 50.7 & 67.3 & 65.9 & 65.4 & 66.2 & 45.9 & 45.0 & 45.4 & 45.4 & 16.4 & 13.6 & 14.2 & 14.7 & 62.1 & 61.6 & 61.7 & 61.8 & 21.6 & 21.3 & 22.5 & 21.8 & $+2.61\%$ \\
    \textit{LESS} & 35.9 & 41.9 & 54.2 & 44.0 & 56.9 & 21.1 & 72.7 & 50.2 & 42.9 & 41.1 & 45.2 & 43.0 & 13.9 & 14.4 & 12.8 & 13.7 & 65.0 & 64.2 & 68.0 & {\underline{65.7}} & 18.0 & 19.4 & 20.9 & 19.5 & $-7.35\%$\\
    \midrule
    \multicolumn{25}{c}{\textbf{\textit{Applying LangGPS (to random selection and three best well-performing baselines marked with *)}}} \\
    \textit{Rand} & 48.8 & 52.0 & 54.9 & 51.9 & 59.1 & 72.2 & 73.2 & 68.2 & 48.8 & 46.2 & 44.2 & 46.4 & 15.4 & 14.8 & 14.5 & 14.9 & 64.7 & 65.0 & 64.9 & {\textbf{64.9}} & 22.0 & 21.7 & 22.4 & 22.0 & $+4.33\%$\\
    & & & & \cellcolor{cyan!15}{(+0.8)} & & & & \cellcolor{cyan!15}{(+0.5)} & & & & \cellcolor{cyan!15}{(+0.03)} & & & & \cellcolor{gray!15}{(-0.1)} & & & & \cellcolor{cyan!15}{(+0.5)} & & & & \cellcolor{gray!15}{(-0.3)} \\
    \textit{KMC} & 50.3 & 53.7 & 53.6 & 52.5 & 78.8 & 70.1 & 72.2 & {\textbf{73.7}} & 43.9 & 46.5 & 45.2 & 45.2 & 13.8 & 14.4 & 13.9 & 14.0 & 62.7 & 62.3 & 63.4 & 62.8 & 21.6 & 23.0 & 23.0 & {\textbf{22.5}} & $+4.32\%$\\
    & & & & \cellcolor{cyan!15}{(+1.7)} & & & & \cellcolor{cyan!15}{(+3.1)} & & & & \cellcolor{cyan!15}{(+0.4)} & & & & \cellcolor{cyan!15}{(+0.4)} & & & & \cellcolor{cyan!15}{(+0.9)} & & & & \cellcolor{cyan!15}{(+0.6)} \\
    \textit{MTLD} & 51.8 & 52.7 & 53.7 & {\textbf{52.7}} & 69.8 & 70.5 & 68.4 & 69.6 & 47.2 & 47.6 & 43.3 & 46.1 & 15.2 & 15.0 & 14.9 & 15.0 & 65.5 & 64.3 & 63.9 & 64.6 & 21.1 & 21.8 & 22.4 & 21.8 & {$\bm{+4.64\%}$}\\
    & & & & \cellcolor{cyan!15}{(+0.9)} & & & & \cellcolor{gray!15}{(-0.3)} & & & & \cellcolor{cyan!15}{(+0.7)} & & & & \cellcolor{cyan!15}{(+0.3)} & & & & \cellcolor{cyan!15}{(+0.7)} & & & & \cellcolor{gray!15}{(-0.4)} \\
    \textit{DSIR} & 49.8 & 54.1 & 50.6 & 51.5 & 62.1 & 67.2 & 69.8 & 66.4 & 48.9 & 50.2 & 46.9 & {\textbf{48.7}} & 15.9 & 14.8 & 15.0 & {\textbf{15.2}}& 64.2 & 63.7 & 64.1 & 64.0 & 21.0 & 22.2 & 22.5 & 21.9 & $+4.62\%$\\
    & & & & \cellcolor{cyan!15}{(+0.8)} & & & & \cellcolor{cyan!15}{(+0.2)} & & & & \cellcolor{cyan!15}{(+3.3)} & & & & \cellcolor{cyan!15}{(+0.5)} & & & & \cellcolor{cyan!15}{(+2.2)} & & & & \cellcolor{cyan!15}{(+0.1)} \\
    \bottomrule
    \end{tabular}
}
\caption{Main results averaged over all languages for each dataset. \textbf{Avg.} denotes the average performance across the 1\%, 3\%, and 5\% training settings. For each model, {\name} is applied on top of both random selection and the three best well-performing baselines (with highest $\Delta$ values) under that model. \colorbox{cyan!15}{Blue cell} indicates better performance than the vanilla baseline under the same training setting, while \colorbox{gray!15}{Gray cell} indicate a performance drop. {\underline{Underline}} numbers indicate the best performance among vanilla baselines, {\textbf{Bold}} numbers indicate the best performance achieved after applying {\name}. $\Delta$ represents the average relative gain or decline (in percentage) of each method across all datasets compared to training on the full dataset.\\
$\Delta(\mathrm{Method})=\frac{1}{|\mathrm{AllDatasets}|}{\sum_{d \in \mathrm{AllDatasets}}{\frac{\mathrm{Result(Method,d)}-\mathrm{Result(Full,d)}}{\mathrm{Result(Full,d)}}}}$}
\label{tab:main}
\end{table*}

%% file: pictures/lang_gains.tex
\definecolor{my-green}{RGB}{100,149,237}  
\definecolor{my-red}{RGB}{221,160,221}  
\definecolor{my-yellow}{RGB}{192,192,192}   

\begin{figure}[t]
\centering
\pgfplotsset{width=1.05\linewidth, height=0.45\linewidth, compat=1.15}
\footnotesize
\begin{tikzpicture}
\footnotesize{
\begin{axis}[
    title={\textit{LLaMA-3.1-8B}},
    title style={yshift=-0.6em, font=\footnotesize},
    at={(0em, 0em)},
    ymajorgrids=true,
    xmajorgrids=true,
    grid style=dashed,
    ybar, 
    bar width=8pt, 
    symbolic x coords={XNLI, XStoryCloze, MMMLU, XQuAD, XLSum}, 
    xtick=data, 
    xticklabel style={font=\fontsize{8.5pt}{10pt}\selectfont, yshift=4pt},
    ymin=-0.5,ymax=11, ytick={0,2,4,6,8,10},
    ylabel={Gains / Declines (\%)}, 
    ytick align=inside,
    legend style={
        at={(0.44,1.2)},
        anchor=south,
        legend columns=2,
        nodes={scale=1, transform shape},
        column sep=0.5em, 
        },
    ]
    \addplot[fill=my-green, draw=none] coordinates {(XNLI,5.36) (XStoryCloze, 7.52) (MMMLU, 0.30) (XQuAD, 0.15) (XLSum, 2.08)};
    \addlegendentry{\text{High-resource Languages}}
    \node[font=\fontsize{6.5pt}{7pt}\selectfont, yshift=3pt, xshift=4pt, anchor=east] at (axis cs:XNLI, 5.36) {5.36\%};
    \node[font=\fontsize{6.5pt}{7pt}\selectfont, yshift=3pt, xshift=4pt, anchor=east] at (axis cs:XStoryCloze, 7.52) {7.52\%};
    \node[font=\fontsize{6.5pt}{7pt}\selectfont, yshift=3pt, xshift=4pt, anchor=east] at (axis cs:MMMLU, 0.30) {0.30\%};
    \node[font=\fontsize{6.5pt}{7pt}\selectfont, yshift=3pt, xshift=3pt, anchor=east] at (axis cs:XQuAD, 0.15) {0.15\%};
    \node[font=\fontsize{6.5pt}{7pt}\selectfont, yshift=3pt, xshift=3pt, anchor=east] at (axis cs:XLSum, 2.08) {2.08\%};

    \addplot[fill=my-red, draw=none] coordinates {(XNLI, 7.27) (XStoryCloze, 8.67) (MMMLU, 9.54) (XQuAD, 0.09) (XLSum, 2.46)};
    \addlegendentry{\text{Low-resource Languages}}
    \node[font=\fontsize{6.5pt}{7pt}\selectfont, yshift=3pt, xshift=-4pt, anchor=west] at (axis cs:XNLI, 7.27) {7.27\%};
    \node[font=\fontsize{6.5pt}{7pt}\selectfont, yshift=3pt, xshift=-4pt, anchor=west] at (axis cs:XStoryCloze, 8.67) {8.67\%};
    \node[font=\fontsize{6.5pt}{7pt}\selectfont, yshift=3pt, xshift=-4pt, anchor=west] at (axis cs:MMMLU, 9.54) {9.54\%};
    \node[font=\fontsize{6.5pt}{7pt}\selectfont, yshift=3pt, xshift=-3pt, anchor=west] at (axis cs:XQuAD, 0.09) {0.09\%};
    \node[font=\fontsize{6.5pt}{7pt}\selectfont, yshift=3pt, xshift=-3pt, anchor=west] at (axis cs:XLSum, 2.46) {2.46\%};
    
    \end{axis}

\begin{axis}[
    title={\textit{Qwen2.5-7B}},
    title style={yshift=-0.6em, font=\footnotesize},
    at={(0em,-10em)},
    ymajorgrids=true,
    xmajorgrids=true,
    grid style=dashed,
    ybar, 
    bar width=8pt, 
    symbolic x coords={XNLI, XStoryCloze, MMMLU, XQuAD, XLSum},
    xtick=data, 
    xticklabel style={font=\fontsize{8.5pt}{10pt}\selectfont, yshift=4pt},
    ymin=-3,ymax=14, ytick={-3,0,3,6,9,12},
    ylabel={Gains / Declines (\%)}, 
    ytick align=inside,
    enlarge x limits=0.1, 
    ]
    \addplot[fill=my-green, draw=none] coordinates {(XNLI,1.83) (XStoryCloze, -1.22) (MMMLU, 0.98) (XQuAD, 1.96) (XLSum, -0.80)};

    \node[font=\fontsize{6.5pt}{7pt}\selectfont, yshift=3pt, xshift=4pt, anchor=east] at (axis cs:XNLI, 1.83) {1.83\%};
    \node[font=\fontsize{6.5pt}{7pt}\selectfont, yshift=-3pt, xshift=4pt, anchor=east] at (axis cs:XStoryCloze, -1.22) {-1.22\%};
    \node[font=\fontsize{6.5pt}{7pt}\selectfont, yshift=3pt, xshift=4pt, anchor=east] at (axis cs:MMMLU, 0.98) {0.98\%};
    \node[font=\fontsize{6.5pt}{7pt}\selectfont, yshift=3pt, xshift=3pt, anchor=east] at (axis cs:XQuAD, 1.96) {1.96\%};
    \node[font=\fontsize{6.5pt}{7pt}\selectfont, yshift=-3pt, xshift=4pt, anchor=east] at (axis cs:XLSum, -0.80) {-0.80\%};

    \addplot[fill=my-red, draw=none] coordinates {(XNLI, 6.71) (XStoryCloze, 5.86) (MMMLU, 11.96) (XQuAD, 0.24) (XLSum, 2.54)};

    \node[font=\fontsize{6.5pt}{7pt}\selectfont, yshift=3pt, xshift=-4pt, anchor=west] at (axis cs:XNLI, 6.71) {6.71\%};
    \node[font=\fontsize{6.5pt}{7pt}\selectfont, yshift=3pt, xshift=-4pt, anchor=west] at (axis cs:XStoryCloze, 5.86) {5.86\%};
    \node[font=\fontsize{6.5pt}{7pt}\selectfont, yshift=3pt, xshift=-4pt, anchor=west] at (axis cs:MMMLU, 11.96) {11.96\%};
    \node[font=\fontsize{6.5pt}{7pt}\selectfont, yshift=3pt, xshift=-3pt, anchor=west] at (axis cs:XQuAD, 0.24) {0.24\%};
    \node[font=\fontsize{6.5pt}{7pt}\selectfont, yshift=3pt, xshift=-4pt, anchor=west] at (axis cs:XLSum, 2.54) {2.54\%};
    
    \end{axis}
    }
    \end{tikzpicture}
    \caption{Average relative gains or declines (in percentage) when applying {\name} on top of \textit{Rand} and the three strongest-performing baselines, shown separately for high-resource and low-resource languages.
    }
    \label{fig:lang_gains}
\end{figure}

%% file: pictures/pre_percent.tex
\definecolor{my-red}{RGB}{255,192,203}
\definecolor{my-green}{RGB}{72,209,204}

\pgfplotsset{compat=1.10}

\begin{figure}[t]
  \footnotesize
  \begin{tikzpicture}[]
      \begin{axis}[
      title={\textit{MMMLU}},
      title style={yshift=-0.5em},
      at={(0em, 0em)},
      ymajorgrids,
      grid style=densely dashed,
      width=0.27\textwidth,
      height=.24\textwidth,
      legend style={
        at={(1.13,1.2)},
        anchor=south,
        legend columns=2,
        nodes={scale=1, transform shape},
        column sep=0.2em,
        },
      xlabel={Pre-selection ratio $\rho$},
      ylabel={Accuracy},
      ylabel style={yshift=-0.5em},xlabel style={yshift=0.05em},
      ymin=36.5,ymax=43, ytick={37, 39, 41, 43},
      xmin=1,xmax=10, xtick={1, 2, 3, 4, 5, 6, 7, 8, 9, 10}, 
      xticklabels={10,20,30,40,50,60,70,80,90,100},
      xticklabel style={font=\scriptsize},
      yticklabel style={font=\scriptsize}
      ]

      \addplot[name path=A, myRed, dashed, line width=1.0pt] coordinates {
        (1,38.1)
        (2,38.1)
        (3,38.1)
        (4,38.1)
        (5,38.1)
        (6,38.1)
        (7,38.1)
        (8,38.1)
        (9,38.1)
        (10,38.1)
      };
      \addlegendentry{\textit{Random-5\%}}

      \addplot[name path=B, myRed,mark=*,thick,mark options={fill=white,draw=myRed,line width=1.0pt}] coordinates {
        (1,36.87)
        (2,39.69)
        (3,39.40)
        (4,39.22)
        (5,39.68)
        (6,39.97)
        (7,38.38)
        (8,37.67)
        (9,39.05)
        (10,38.1)
      };
      \addlegendentry{{\name}+\textit{Random-5\%}}

      \addplot[name path=C, my-green, dashed, line width=1.0pt] coordinates {
        (1,40.8)
        (2,40.8)
        (3,40.8)
        (4,40.8)
        (5,40.8)
        (6,40.8)
        (7,40.8)
        (8,40.8)
        (9,40.8)
        (10,40.8)
        };
      \addlegendentry{\textit{LESS-5\%}}

      \addplot[name path=D, my-green,mark=*,thick,mark options={fill=white,draw=my-green,line width=1.0pt}] coordinates {
        (1,42.0)
        (2,42.2)
        (3,41.4)
        (4,41.7)
        (5,42.2)
        (6,41.3)
        (7,42.3)
        (8,41.9)
        (9,40.4)
        (10,40.8)
      };
      \addlegendentry{{\name}+\textit{LESS-5\%}}

      \addplot[my-red!20] fill between[of=A and B];
      
      \addplot [my-green!20] fill between [of=C and D];
      \end{axis}

      \begin{axis}[
      title={\textit{XLSum}},
      title style={yshift=-0.55em},
      at={(13em,0em)},
      ymajorgrids,
      grid style=densely dashed,
      width=0.27\textwidth,
      height=.24\textwidth,
      xlabel={Pre-selection ratio $\rho$},
      ylabel={Rouge-L},
      ylabel style={yshift=-0.5em},xlabel style={yshift=0.05em},
      ymin=22.3,ymax=24.3,ytick={22,23,24},
      xmin=1,xmax=10,xtick={1, 2, 3, 4, 5, 6, 7, 8, 9, 10}, 
      xticklabels={10,20,30,40,50,60,70,80,90,100},
      xticklabel style={font=\scriptsize},
      yticklabel style={font=\scriptsize}
      ]

      \addplot[name path=A, myRed, dashed, line width=1.0pt] coordinates {
        (1,22.5)
        (2,22.5)
        (3,22.5)
        (4,22.5)
        (5,22.5)
        (6,22.5)
        (7,22.5)
        (8,22.5)
        (9,22.5)
        (10,22.5)
      };

      \addplot[name path=B, myRed,mark=*,thick,mark options={fill=white,draw=myRed,line width=1.0pt}] coordinates {
        (1,23.10)
        (2,22.9)
        (3,23.45)
        (4,23.54)
        (5,23.13)
        (6,22.91)
        (7,22.77)
        (8,22.68)
        (9,22.55)
        (10,22.5)
      };

      \addplot[name path=C, my-green, dashed, line width=1.0pt] coordinates {
        (1,22.8)
        (2,22.8)
        (3,22.8)
        (4,22.8)
        (5,22.8)
        (6,22.8)
        (7,22.8)
        (8,22.8)
        (9,22.8)
        (10,22.8)
        };

      \addplot[name path=D, my-green,mark=*,thick,mark options={fill=white,draw=my-green,line width=1.0pt}] coordinates {
        (1,24.1)
        (2,23.4)
        (3,23.2)
        (4,22.9)
        (5,23.3)
        (6,23.0)
        (7,22.9)
        (8,23.1)
        (9,22.7)
        (10,22.8)
      };
      \addplot[my-red!20] fill between[of=A and B];
      \addplot[my-green!20] fill between [of=C and D];
      \end{axis}
    \end{tikzpicture}
  \caption{Effect of the pre-selection ratio $\rho$ on performance (\textit{MMMLU} and \textit{XLSum}, 5\% setting, \textit{LLaMA-3.1-8B}). We vary $\rho$ from 10\% to 100\% and report results using both \textit{Rand} and \textit{LESS} as downstream selectors.}
  \label{fig:pre_percent}
\end{figure}

%% file: pictures/translation.tex
\definecolor{my-blue}{RGB}{72,209,204}
\definecolor{my-red}{RGB}{220, 72, 72}
\definecolor{my-yellow}{RGB}{240, 180, 40}


\begin{figure}[t]
  \centering
  \begin{tikzpicture}[]
      \scriptsize{
      \begin{axis}[
      title={Avg. Performance on \textit{FLORES+}: $X \rightarrow En$ \textit{(X=De,Fr,Zh)}},
      title style={yshift=-0.8em, font=\footnotesize},
      at={(12em,0em)}, 
      anchor=north west,  
      ymajorgrids,
      grid style=densely dashed,
      width=0.48\textwidth,
      height=.25\textwidth,
      legend style={
        at={(0.48,1.2)},
        anchor=south,
        legend columns=3,
        nodes={scale=0.8, transform shape},
        column sep=0.5em,
        font=\normalsize,
        },
      xlabel={\footnotesize{Training Data Scale (\# of Samples)}},
      ylabel={\footnotesize{COMET}},
      ylabel style={yshift=0.25em},xlabel style={yshift=0.05em},
      yticklabel style={
            /pgf/number format/fixed, 
            /pgf/number format/precision=2  
        },
      ymin=40,ymax=68, ytick={40,45,50,55,60,65},
      xmin=1,xmax=18.5, xtick={1, 2, 3, 4, 5, 6, 7, 8, 9, 10, 11, 12.5, 14, 15.5, 17, 18.5}, 
      xticklabels={0, 50, 100, 150, 200, 250, 300, 350, 400, 450, 500, 1000, 2000, 3000, 4000, 5000},
      ]

      \fill[my-red, opacity=0.04] (axis cs:1,40) rectangle (axis cs:5,68); 
      \fill[my-blue, opacity=0.04] (axis cs:5,40) rectangle (axis cs:17,68); 
      \fill[my-red, opacity=0.04] (axis cs:17,40) rectangle (axis cs:18.5,68); 

      \draw[dashed, thick, opacity=0.8] (axis cs:5,40) -- (axis cs:5,68);
      \draw[dashed, thick, opacity=0.8] (axis cs:17,40) -- (axis cs:17,68);

      \addplot[my-blue,smooth,mark=*,thick,mark options={fill=white,draw=my-blue,line width=1.0pt}] coordinates {
        (1,62.48)
        (2,62.78)
        (3,62.38)
        (4,61.81)
        (5,61.83)
        (6,62.28)
        (7,58.68)
        (8,58.29)
        (9,46.78)
        (10,58.84)
        (11,61.65)
        (12.5,44.87)
        (14,62.92)
        (15.5,62.94)
        (17,62.86)
        (18.5,62.95)
      };
      \addlegendentry{High-Separability}

      \addplot[my-red,smooth,mark=*,thick,mark options={fill=white,draw=my-red,line width=1.0pt}] coordinates {
        (1,62.48)
        (2,61.35)
        (3,62.52)
        (4,62.57)
        (5,62.46)
        (6,49.82)
        (7,61.36)
        (8,52.13)
        (9,64.89)
        (10,59.15)
        (11,66.06)
        (12.5,58.27)
        (14,62.91)
        (15.5,65.29)
        (17,63.56)
        (18.5,63.04)
      };
      \addlegendentry{Low-Separability}

      \addplot[my-yellow,smooth,mark=*,thick,mark options={fill=white,draw=my-yellow,line width=1.0pt}] coordinates {
        (1,62.48)
        (2,62.64)
        (3,62.39)
        (4,62.68)
        (5,62.76)
        (6,56.83)
        (7,43.05)
        (8,61.35)
        (9,42.49)
        (10,64.35)
        (11,56.02)
        (12.5,63.34)
        (14,62.82)
        (15.5,63.34)
        (17,62.82)
        (18.5,63.10)
      };
      \addlegendentry{Random}

      \end{axis}
     }
    \end{tikzpicture}

    \begin{tikzpicture}[]
      \scriptsize{
      \begin{axis}[
      title={Avg. Performance on \textit{FLORES+}: $En \rightarrow X$ \textit{(X=De,Fr,Zh)}},
      title style={yshift=-0.8em, font=\footnotesize},
      at={(12em,55em)}, 
      anchor=north west,  
      ymajorgrids,
      grid style=densely dashed,
      width=0.48\textwidth,
      height=.25\textwidth,
      xlabel={\footnotesize{Training Data Scale (\# of Samples)}},
      ylabel={\footnotesize{COMET}},
      ylabel style={yshift=0.25em},xlabel style={yshift=0.05em},
      yticklabel style={
            /pgf/number format/fixed, 
            /pgf/number format/precision=2  
        },
      ymin=50,ymax=84, ytick={50,55,60,65,70,75,80},
      xmin=1,xmax=18.5, xtick={1, 2, 3, 4, 5, 6, 7, 8, 9, 10, 11, 12.5, 14, 15.5, 17, 18.5}, 
      xticklabels={0, 50, 100, 150, 200, 250, 300, 350, 400, 450, 500, 1000, 2000, 3000, 4000, 5000},
      ]
      \fill[my-blue, opacity=0.04] (axis cs:1,50) rectangle (axis cs:14,84); 
      \fill[my-red, opacity=0.04] (axis cs:14,50) rectangle (axis cs:18.5,84); 
      
      \draw[dashed, thick, opacity=0.8] (axis cs:14,50) -- (axis cs:14,84);
      
      \addplot[my-blue,smooth,mark=*,thick,mark options={fill=white,draw=my-blue,line width=1.0pt}] coordinates {
        (1,51.37)
        (2,69.14)
        (3,69.22)
        (4,73.72)
        (5,75.88)
        (6,75.07)
        (7,69.09)
        (8,68.89)
        (9,60.27)
        (10,66.32)
        (11,68.67)
        (12.5,59.64)
        (14,81.05)
        (15.5,80.72)
        (17,80.18)
        (18.5,80.42)
      };

      \addplot[my-red,smooth,mark=*,thick,mark options={fill=white,draw=my-red,line width=1.0pt}] coordinates {
        (1,51.37)
        (2,55.36)
        (3,67.36)
        (4,77.62)
        (5,79.01)
        (6,67.98)
        (7,66.38)
        (8,66.82)
        (9,67.66)
        (10,64.94)
        (11,68.74)
        (12.5,69.69)
        (14,80.85)
        (15.5,80.27)
        (17,79.74)
        (18.5,79.92)
      };

      \addplot[my-yellow,smooth,mark=*,thick,mark options={fill=white,draw=my-yellow,line width=1.0pt}] coordinates {
        (1,51.37)
        (2,67.05)
        (3,73.58)
        (4,76.99)
        (5,79.59)
        (6,65.69)
        (7,65.58)
        (8,66.95)
        (9,61.02)
        (10,66.29)
        (11,67.72)
        (12.5,80.21)
        (14,80.72)
        (15.5,80.21)
        (17,80.72)
        (18.5,80.39)
      };

      \end{axis}
     }
    \end{tikzpicture}
  \caption{
    Average translation performance of \textit{LLaMA-3.1-8B} on \textit{FLORES+} ($X \rightarrow En$ and $En \rightarrow X$, where $X$ = De, Fr, Zh) under varying training sizes, comparing models trained on top separability-scoring, bottom separability-scoring, and randomly selected samples.
    }
  \label{fig:translation}
\end{figure}

%% file: tables/curriculum.tex
\begin{table}[t]
\small
\setlength{\tabcolsep}{3pt}
\renewcommand{\arraystretch}{1.1} 
  \centering
  \resizebox{\columnwidth}{!}{
    \begin{tabular}{lcccccc}
    \toprule
    \multirow{2}{*}{Datasets} & \multirow{2}{*}{XNLI} & \multirow{2}{*}{\shortstack{XStory\\Cloze}} & \multirow{2}{*}{MMMLU} & \multirow{2}{*}{MKQA} & \multirow{2}{*}{XQuAD} & \multirow{2}{*}{XLSum} \\
    \\
    \midrule
    Strategy & \multicolumn{6}{c}{\textit{Model: LLaMA-3.1-8B}} \\
    \midrule
    \textit{Random} & 33.9 & 56.8 & \textbf{36.8} & \underline{11.1} & 59.8 & \underline{22.1} \\
    \textit{Ascending} & \textbf{37.9} & 67.5 & 32.5 & 11.0 & 58.7 & 21.3 \\
    \textit{Descending} & 33.1 & 63.4 & 33.2 & 10.8 & \underline{61.0} & \textbf{22.2} \\
    \textit{Balanced} & \underline{36.8} & \textbf{68.8} & \underline{34.4} & \textbf{12.1} & \textbf{61.0} & 22.0 \\
    \toprule
    Strategy & \multicolumn{6}{c}{\textit{Model: Qwen2.5-7B}} \\
    \midrule
    \textit{Random} & \underline{45.3} & 74.3 & \underline{48.6} & 11.8 & 61.4 & 21.5 \\
    \textit{Ascending} & 36.4 & \textbf{76.1} & 44.6 & 11.5 & \underline{61.8} & 21.5 \\
    \textit{Descending} & 44.5 & 61.6 & 45.1 & \textbf{12.2} & 61.7 & \underline{21.9} \\
    \textit{Balanced} & \textbf{50.8} & \underline{75.9} & \textbf{49.7} & \underline{11.9} & \textbf{62.7} & \textbf{22.3} \\
    \bottomrule
    \end{tabular}
  }
  \caption{Results of multilingual curriculum learning. We compare the three curriculum strategies—\textit{Ascending}, \textit{Descending}, and \textit{Balanced}—with \textit{Randomly Shuffled}.
  }
  \label{tab:curriculum}
\end{table}

%% file: tables/time.tex
\begin{table*}[t]
    \centering
    \renewcommand{\arraystretch}{1} 
    \small
    \begin{tabularx}{0.95\textwidth}{>{\centering\arraybackslash}p{1.8cm}*{6}{>{\centering\arraybackslash}X}}
    \toprule
    \multirow{2}[2]{*}{\textbf{Composition}} & \multicolumn{2}{c}{\textbf{Getting Representations}} & \multicolumn{2}{c}{\textbf{Clustering}} & \multicolumn{2}{c}{\textbf{Data Selection}} \\
    \cmidrule(lr){2-7}
    & {\textbf{Complexity}} &  {\textbf{Actual}} & {\textbf{Complexity}} & {\textbf{Actual}} & {\textbf{Complexity}} & {\textbf{Actual}} \\ 
    \cmidrule(lr){1-3} \cmidrule(lr){4-5} \cmidrule(lr){6-7}
    {\textit{KMC}} & $\mathcal{O}(\vert \mathcal{D} \vert)$ & $10$ Mins & $\mathcal{O}(\vert \mathcal{D} \vert \cdot K)$ & $1$ Min & $\mathcal{O}(\vert \mathcal{D} \vert)$ & $1$ Min  \\
    \end{tabularx}
    \begin{tabularx}{0.95\textwidth}{>{\centering\arraybackslash}p{1.8cm}*{4}{>{\centering\arraybackslash}X}}
    \toprule
    \multirow{2}[2]{*}{\textbf{Composition}} & \multicolumn{2}{c}{\textbf{MTLD Scoring}}  & \multicolumn{2}{c}{\textbf{Data Selection}} \\
    \cmidrule(lr){2-5}
    & {\textbf{Complexity}} &  {\textbf{Actual}} & {\textbf{Complexity}} & {\textbf{Actual}} \\ 
    \cmidrule(lr){1-3} \cmidrule(lr){4-5}
    {\textit{MTLD}} & $\mathcal{O}(\vert \mathcal{D} \vert)$ & $10$ Mins & $\mathcal{O}(\vert \mathcal{D} \vert \cdot \rm{log} \vert \mathcal{D} \vert)$ & $1$ Min \\
    \end{tabularx}
    \begin{tabularx}{0.95\textwidth}{>{\centering\arraybackslash}p{1.8cm}*{4}{>{\centering\arraybackslash}X}}
    \toprule
    \multirow{2}[2]{*}{\textbf{Composition}} & \multicolumn{2}{c}{\textbf{UniEval Scoring}}  & \multicolumn{2}{c}{\textbf{Data Selection}} \\
    \cmidrule(lr){2-5}
    & {\textbf{Complexity}} &  {\textbf{Actual}} & {\textbf{Complexity}} & {\textbf{Actual}} \\ 
    \cmidrule(lr){1-3} \cmidrule(lr){4-5}
    {\textit{Nat}/\textit{Coh}/\textit{Und}} & $\mathcal{O}(\vert \mathcal{D} \vert)$ & $2$ Hours & $\mathcal{O}(\vert \mathcal{D} \vert \cdot \rm{log} \vert \mathcal{D} \vert)$ & $1$ Min \\
    \end{tabularx}
    \begin{tabularx}{0.95\textwidth}{>{\centering\arraybackslash}p{1.8cm}*{4}{>{\centering\arraybackslash}X}}
    \toprule
    \multirow{2}[2]{*}{\textbf{Composition}} & \multicolumn{2}{c}{\textbf{Importance Weights Computation \& Data Selection}} \\
    \cmidrule(lr){2-3}
    & {\textbf{Complexity}} &  {\textbf{Actual}}  \\ 
    \cmidrule(lr){1-3}
    {\textit{DSIR}} & - & $3$ Mins  \\
    \end{tabularx}
    \begin{tabularx}{0.95\textwidth}{>{\centering\arraybackslash}p{1.8cm}*{6}{>{\centering\arraybackslash}X}}
    \toprule
    \multirow{2}[2]{*}{\textbf{Composition}} & \multicolumn{2}{c}{\textbf{Warmup LoRA Training}}& \multicolumn{2}{c}{\textbf{Gradient Features Computation}} & \multicolumn{2}{c}{\textbf{Data Selection}} \\
    \cmidrule(lr){2-7}
    & \textbf{Complexity} & \textbf{Actual} & \textbf{Complexity} & \textbf{Actual} & \textbf{Complexity} & \textbf{Actual} \\ 
    \cmidrule(lr){1-3} \cmidrule(lr){4-5} \cmidrule(lr){6-7} 
    \textit{LESS} & $\mathcal{O}(K)$ & $2$ Hours & $\mathcal{O}(\mathcal{D})$ & $6$ Hours & $\mathcal{O}(\vert \mathcal{D} \vert \cdot \vert \mathcal{D}_{tgt} \vert)$ & $1$ Min \\
    \end{tabularx}
    \begin{tabularx}{0.95\textwidth}{>{\centering\arraybackslash}p{1.8cm}*{6}{>{\centering\arraybackslash}X}}
    \toprule
    \multirow{2}[2]{*}{\textbf{Composition}} & \multicolumn{2}{c}{\textbf{Getting Representations}}& \multicolumn{2}{c}{\textbf{Silhouette Scoring}} & \multicolumn{2}{c}{\textbf{Data Selection}} \\
    \cmidrule(lr){2-7}
    & \textbf{Complexity} & \textbf{Actual} & \textbf{Complexity} & \textbf{Actual} & \textbf{Complexity} & \textbf{Actual} \\ 
    \cmidrule(lr){1-3} \cmidrule(lr){4-5} \cmidrule(lr){6-7} 
    {\name} & $\mathcal{O}(\vert \mathcal{D} \vert)$ & $2$ Hours (fp16) & $\mathcal{O}(\vert \mathcal{D} \vert^{2})$ & $1$ Min & $\mathcal{O}(\vert \mathcal{D} \vert \cdot \rm{log} \vert \mathcal{D} \vert)$ & $1$ Min \\
    \bottomrule
    \end{tabularx}
    \caption{The computational cost of {\name} and other baselines on full training corpus (measured as \textbf{single} A100 GPU hours, $\vert \mathcal{D} \vert=97696$, $K=5\% \times \vert \mathcal{D} \vert=4885$ and $\mathcal{D}_{tgt}$ represents the target set for the target-dependent baselines). Getting Representations is the most costly step in {\name}.}
    \label{tab:cost}
\end{table*}

%% file: tables/prompts.tex
\definecolor{given}{RGB}{197,217,197}
\definecolor{response}{RGB}{176,224,230}

\begin{table*}[ht]
    \small
    \setlength\tabcolsep{3pt}
        \begin{tabular}{p{\textwidth}}
        \toprule
        Prompt for \textit{XNLI} (English version) \\
        \midrule
        Premise: \colorbox{given}{\{premise\}} \\
        Hypothesis: \colorbox{given}{\{hypothesis\}} \\
        \\
        What do you think is the relationship between the premise and the hypothesis? \\
        \\
        (1) Entail \\
        (2) Neutral \\
        (3) Contradict \\
        \\
        If you have to choose one of these options, your answer would be:
        \colorbox{response}{\{response\}} \\
        \toprule
        Prompt for \textit{XStoryCloze} (English version) \\
        \midrule
        \colorbox{given}{\{story\}} \\
        (1) \colorbox{given}{\{first possible continuation of the story\}} \\
        (2) \colorbox{given}{\{second possible continuation of the story\}} \\
        \\
        Which of the two options is more likely to be the ending of the given story. \\
        Your answer:
        \colorbox{response}{\{response\}} \\
        \toprule
        Prompt for \textit{MMMLU} (English version) \\
        \midrule
        Please answer the following multiple choice question. \\
        \colorbox{given}{\{question\}} \\        
        (A) \colorbox{given}{\{option 1\}} \\
        (B) \colorbox{given}{\{option 2\}} \\
        (C) \colorbox{given}{\{option 3\}} \\
        (D) \colorbox{given}{\{option 4\}} \\
        \\
        If you have to choose one of these options, your answer would be:
        \colorbox{response}{\{response\}} \\
        \toprule
        Prompt for \textit{MKQA} (English version) \\
        \midrule
        Please answer the following question. \\
        \\
        \colorbox{given}{\{question\}} \\
        \\
        Your answer: \colorbox{response}{\{response\}} \\
        \toprule
        Prompt for \textit{XQuAD} (English version) \\
        \midrule
        Please answer these questions only based on the given context. \\
        \\
        Context: \colorbox{given}{\{context\}} \\
        Question: \colorbox{given}{\{question\}} \\
        \\
        Your answer: \colorbox{response}{\{response\}} \\
        \toprule
        Prompt for \textit{XLSum} (English version) \\
        \midrule
        Please summarize the following content into one sentence. \\
        \colorbox{given}{\{content\}} \\
        \colorbox{response}{\{response\}} \\
        \toprule
        \end{tabular}
    \caption{The prompts used for \textit{XNLI}, \textit{XStoryCloze}, \textit{MMMLU}, \textit{MKQA}, \textit{XQuAD} and \textit{XLSum}.}
    \label{tab:prompts}
\end{table*}

%% file: tables/main_xnli.tex
\begin{table*}[p]
\centering
\setlength{\aboverulesep}{1pt} 
\setlength{\belowrulesep}{1pt} 
\renewcommand{\arraystretch}{1.1} 
\setlength{\tabcolsep}{1.5pt}

\resizebox{\linewidth}{!}{
\begin{tabular}{lcccccccccc|c|cccccccccc|c|cccccccccc|c}
    \toprule
    \multirow{3}[3]{*}{\textbf{Method}} & \multicolumn{33}{c}{\textbf{\textit{Model: LLaMA-3.1-8B}}} \\
    \cmidrule(lr){2-34}
    & \multicolumn{33}{c}{\textbf{Dataset: XNLI}} \\
    \cmidrule(lr){2-34}
     & \multicolumn{11}{c}{1\%} & \multicolumn{11}{c}{3\%} & \multicolumn{11}{c}{5\%} \\
    \midrule
    Langs & en & ar & el & es & fr & hi & sw & tr & vi & zh & {\textbf{Avg.}} & en & ar & el & es & fr & hi & sw & tr & vi & zh & {\textbf{Avg.}} &  en & ar & el & es & fr & hi & sw & tr & vi & zh  & {\textbf{Avg.}}\\
    \midrule
    \textit{Rand} & 35.4 & 10.0 & 23.1 & 34.6 & 30.0 & 10.6 & 29.3 & 31.4 & 36.2 & 10.5 & 25.1 & 35.1 & 30.0 & 30.7 & 35.7 & 36.6 & 30.7 & 30.0 & 31.2 & 39.7 & 30.4 & 33.0 & 40.6 & 31.2 & 33.2 & 40.9 & 39.6 & 30.7 & 32.4 & 35.1 & 37.3 & 32.5 & 35.4 \\
    \midrule
    \multicolumn{34}{c}{\textbf{\textit{Feature-based Baselines}}} \\
    
    \textit{KMC*} & 28.8 & 29.7 & 26.5 & 30.7 & 17.2 & 18.5 & 29.5 & 23.7 & 36.2 & 1.7 & 24.3 & 37.8 & 32.7 & 31.0 & 37.3 & 35.2 & 31.4 & 34.9 & 36.4 & 41.9 & 36.4 & 35.5 & 19.9 & 31.7 & 30.4 & 44.8 & 40.9 & 33.7 & 30.6 & 36.7 & 37.2 & 34.8 & 34.1 \\
    \textit{MTLD} & 33.8 & 29.1 & 31.6 & 29.6 & 35.4 & 33.2 & 30.5 & 4.7 & 30.8 & 29.0 & 28.8 & 36.5 & 30.5 & 37.2 & 36.6 & 37.6 & 32.2 & 36.8 & 36.7 & 33.2 & 31.4 & 34.9 & 36.9 & 31.0 & 33.5 & 40.9 & 38.9 & 30.5 & 32.1 & 36.7 & 34.1 & 29.5 & 34.4 \\
\bottomrule
    \end{tabular}
}
\resizebox{\linewidth}{!}{
    \begin{tabular}{lcccccccccc|c|cccccccccc|c|cccccccccc|c}
    \toprule
    \multirow{3}[3]{*}{\textbf{Method}} & \multicolumn{33}{c}{\textbf{\textit{Model: Qwen2.5-7B}}} \\
    \cmidrule(lr){2-34}
    & \multicolumn{33}{c}{\textbf{Dataset: XNLI}} \\
    \cmidrule(lr){2-34}
     & \multicolumn{11}{c}{1\%} & \multicolumn{11}{c}{3\%} & \multicolumn{11}{c}{5\%} \\
    \midrule
    Langs & en & ar & el & es & fr & hi & sw & tr & vi & zh & {\textbf{Avg.}} & en & ar & el & es & fr & hi & sw & tr & vi & zh & {\textbf{Avg.}} &  en & ar & el & es & fr & hi & sw & tr & vi & zh  & {\textbf{Avg.}}\\
    \midrule
    \textit{Rand} & 81.3 & 31.2 & 26.0 & 74.5 & 31.7 & 32.1 & 24.4 & 49.0 & 54.8 & 50.8 & 45.6 & 87.1 & 43.6 & 39.4 & 78.7 & 63.0 & 39.6 & 19.8 & 48.7 & 55.0 & 54.8 & 53.0 & 87.1 & 53.0 & 41.8 & 78.6 & 62.7 & 41.4 & 25.9 & 46.5 & 52.5 & 57.3 & 54.7\\
    \midrule
    \multicolumn{34}{c}{\textbf{\textit{Feature-based Baselines}}} \\
    \textit{KMC*} & 84.3 & 50.0 & 36.8 & 75.4 & 51.6 & 32.9 & 29.7 & 47.4 & 52.4 & 50.9 & 51.1 & 79.9 & 54.8 & 44.3 & 75.2 & 58.7 & 39.8 & 29.7 & 42.9 & 17.0 & 56.0 & 49.8 & 82.5 & 46.8 & 34.9 & 71.5 & 57.4 & 40.4 & 27.0 & 43.8 & 53.4 & 55.4 & 51.3 \\
    \textit{MTLD*} & 88.6 & 48.8 & 44.9 & 83.0 & 45.0 & 35.3 & 4.6 & 46.7 & 50.2 & 53.6 & 50.1 & 88.3 & 50.5 & 40.1 & 78.4 & 61.0 & 37.7 & 28.6 & 50.3 & 51.1 & 53.4 & 53.9 & 84.0 & 49.4 & 35.8 & 75.5 & 55.4 & 36.1 & 26.6 & 47.7 & 50.7 & 53.4 & 51.5 \\
\bottomrule
    \end{tabular}
}
\caption{The detailed performance results of different language subset on \textit{XNLI} dataset.}
\label{tab:main_xnli}
\end{table*}

%% file: tables/main_xstorycloze.tex
\begin{table*}[p]
\centering
\setlength{\aboverulesep}{1pt} 
\setlength{\belowrulesep}{1pt} 
\renewcommand{\arraystretch}{1.1} 
\setlength{\tabcolsep}{1.5pt}

\resizebox{\linewidth}{!}{
    \begin{tabular}{lcccccccccc|c|cccccccccc|c|cccccccccc|c}
    \toprule
    \multirow{3}[3]{*}{\textbf{Method}} & \multicolumn{33}{c}{\textbf{\textit{Model: LLaMA-3.1-8B}}} \\
    \cmidrule(lr){2-34}
    & \multicolumn{33}{c}{\textbf{Dataset: XStoryCloze}} \\
    \cmidrule(lr){2-34}
     & \multicolumn{11}{c}{1\%} & \multicolumn{11}{c}{3\%} & \multicolumn{11}{c}{5\%} \\
    \midrule
    Langs & en & ar & es & eu & hi & id & ru & sw & te & zh & {\textbf{Avg.}} & en & ar & es & eu & hi & id & ru & sw & te & zh & {\textbf{Avg.}} & en & ar & es & eu & hi & id & ru & sw & te & zh & {\textbf{Avg.}}\\
    \midrule
    \textit{Rand} & 51.1 & 48.6 & 54.1 & 26.9 & 55.9 & 46.3 & 49.7 & 28.0 & 19.6 & 56.7 & 43.7 & 67.3 & 60.3 & 63.7 & 49.2 & 68.1 & 64.2 & 74.1 & 44.7 & 45.1 & 77.6 & 61.4 & 80.9 & 43.9 & 57.1 & 52.1 & 68.0 & 74.2 & 73.8 & 44.4 & 38.7 & 77.6 & 61.1 \\
    \midrule
    \multicolumn{34}{c}{\textbf{\textit{Feature-based Baselines}}} \\
    
    \textit{KMC*} & 41.6 & 18.9 & 49.8 & 24.2 & 64.3 & 55.1 & 25.6 & 19.4 & 59.5 & 83.0 & 44.1 & 67.8 & 61.6 & 73.6 & 52.1 & 69.3 & 72.0 & 69.3 & 37.8 & 55.1 & 78.0 & 63.7 & 88.6 & 76.5 & 55.5 & 53.1 & 63.5 & 79.0 & 53.3 & 49.8 & 54.1 & 82.7 & 65.6 \\
    \textit{MTLD} & 38.0 & 6.0 & 36.3 & 24.0 & 60.0 & 65.1 & 32.4 & 19.0 & 0.0 & 62.3 & 34.3 & 82.0 & 10.1 & 68.2 & 47.3 & 57.3 & 72.9 & 56.3 & 48.6 & 11.9 & 74.5 & 52.9 & 67.2 & 14.4 & 56.7 & 52.0 & 21.0 & 69.6 & 44.3 & 37.9 & 27.3 & 80.7 & 47.1 \\
\bottomrule
    \end{tabular}
}
\resizebox{\linewidth}{!}{
    \begin{tabular}{lcccccccccc|c|cccccccccc|c|cccccccccc|c}
    \toprule
    \multirow{3}[3]{*}{\textbf{Method}} & \multicolumn{33}{c}{\textbf{\textit{Model: Qwen2.5-7B}}} \\
    \cmidrule(lr){2-34}
    & \multicolumn{33}{c}{\textbf{Dataset: XStoryCloze}} \\
    \cmidrule(lr){2-34}
     & \multicolumn{11}{c}{1\%} & \multicolumn{11}{c}{3\%} & \multicolumn{11}{c}{5\%} \\
    \midrule
    Langs & en & ar & el & es & fr & hi & sw & tr & vi & zh & {\textbf{Avg.}} & en & ar & el & es & fr & hi & sw & tr & vi & zh & {\textbf{Avg.}} &  en & ar & el & es & fr & hi & sw & tr & vi & zh  & {\textbf{Avg.}}\\
    \midrule
    \textit{Rand} & 60.6 & 50.6 & 41.6 & 57.8 & 61.4 & 55.4 & 54.8 & 31.4 & 55.2 & 73.5 & 54.2 & 90.2 & 88.6 & 90.8 & 58.1 & 76.7 & 85.3 & 90.2 & 50.4 & 50.9 & 78.5 & 76.0 & 83.5 & 85.9 & 90.1 & 52.2 & 74.3 & 83.9 & 89.6 & 43.4 & 50.4 & 74.2 & 72.7 \\
    \midrule
    \multicolumn{34}{c}{\textbf{\textit{Feature-based Baselines}}} \\
    \textit{KMC*} & 90.5 & 84.1 & 91.7 & 58.2 & 78.7 & 85.8 & 86.6 & 27.1 & 56.5 & 83.7 & 74.3 & 78.1 & 81.1 & 57.4 & 50.6 & 66.0 & 76.5 & 61.9 & 54.2 & 39.7 & 78.6 & 64.4 & 84.5 & 86.0 & 90.5 & 56.3 & 74.0 & 76.4 & 74.3 & 52.5 & 58.9 & 76.4 & 73.0 \\
    \textit{MTLD*} & 83.2 & 58.0 & 80.7 & 52.9 & 65.5 & 80.1 & 82.4 & 31.4 & 54.3 & 53.3 & 64.2 & 84.3 & 85.7 & 83.8 & 57.0 & 74.7 & 86.8 & 87.2 & 49.1 & 50.6 & 76.3 & 73.6 & 89.0 & 83.5 & 80.3 & 56.9 & 74.1 & 86.7 & 70.9 & 50.1 & 47.3 & 80.8 & 72.0 \\
\bottomrule
    \end{tabular}
}
\caption{The detailed performance results of different language subset on \textit{XStoryCloze} dataset.}
\label{tab:main_xstorycloze}
\end{table*}

%% file: tables/main_mmmlu.tex
\begin{table*}[p]
\centering
\setlength{\aboverulesep}{1pt} 
\setlength{\belowrulesep}{1pt} 
\renewcommand{\arraystretch}{1.1} 
\setlength{\tabcolsep}{1.5pt}

\resizebox{\linewidth}{!}{
    \begin{tabular}{lcccccccccc|c|cccccccccc|c|cccccccccc|c}
    \toprule
    \multirow{3}[3]{*}{\textbf{Method}} & \multicolumn{33}{c}{\textbf{\textit{Model: LLaMA-3.1-8B}}} \\
    \cmidrule(lr){2-34}
    & \multicolumn{33}{c}{\textbf{Dataset: MMMLU}} \\
    \cmidrule(lr){2-34}
     & \multicolumn{11}{c}{1\%} & \multicolumn{11}{c}{3\%} & \multicolumn{11}{c}{5\%} \\
    \midrule
    Langs & en & ar & bn & de & es & fr & hi & sw & yo & zh & {\textbf{Avg.}} & en & ar & bn & de & es & fr & hi & sw & yo & zh & {\textbf{Avg.}} & en & ar & bn & de & es & fr & hi & sw & yo & zh & {\textbf{Avg.}}\\
    \midrule
    \textit{Rand} & 52.5 & 31.9 & 29.6 & 42.4 & 45.2 & 40.8 & 26.5 & 25.5 & 13.5 & 31.3 & 34.2 & 54.7 & 39.2 & 29.0 & 44.8 & 46.6 & 46.8 & 33.2 & 26.8 & 17.2 & 43.9 & 37.4 & 56.6 & 38.3 & 28.5 & 46.4 & 50.0 & 45.1 & 35.0 & 28.6 & 21.1 & 43.9 & 38.1 \\
    \midrule
    \multicolumn{34}{c}{\textbf{\textit{Feature-based Baselines}}} \\
    
    \textit{KMC*} & 58.4 & 40.8 & 22.1 & 45.9 & 48.9 & 41.3 & 35.3 & 31.9 & 13.4 & 43.0 & 38.1 & 56.5 & 38.9 & 29.2 & 47.3 & 52.4 & 46.2 & 33.4 & 26.9 & 22.7 & 42.3 & 39.6 & 57.7 & 40.1 & 27.2 & 44.5 & 47.7 & 39.2 & 34.6 & 21.8 & 13.7 & 45.3 & 37.2 \\
    \textit{MTLD} & 43.6 & 7.8 & 2.2 & 39.1 & 38.5 & 25.1 & 28.2 & 16.6 & 13.1 & 36.6 & 25.1 & 53.4 & 33.8 & 6.7 & 43.5 & 42.6 & 34.0 & 31.1 & 24.9 & 15.9 & 39.2 & 32.5 & 50.0 & 34.6 & 1.9 & 42.9 & 46.2 & 32.8 & 24.2 & 20.3 & 6.8 & 38.8 & 29.9 \\
\bottomrule
    \end{tabular}
}
\resizebox{\linewidth}{!}{
    \begin{tabular}{lcccccccccc|c|cccccccccc|c|cccccccccc|c}
    \toprule
    \multirow{3}[3]{*}{\textbf{Method}} & \multicolumn{33}{c}{\textbf{\textit{Model: Qwen2.5-7B}}} \\
    \cmidrule(lr){2-34}
    & \multicolumn{33}{c}{\textbf{Dataset: MMMLU}} \\
    \cmidrule(lr){2-34}
     & \multicolumn{11}{c}{1\%} & \multicolumn{11}{c}{3\%} & \multicolumn{11}{c}{5\%} \\
    \midrule
    Langs & en & ar & el & es & fr & hi & sw & tr & vi & zh & {\textbf{Avg.}} & en & ar & el & es & fr & hi & sw & tr & vi & zh & {\textbf{Avg.}} &  en & ar & el & es & fr & hi & sw & tr & vi & zh  & {\textbf{Avg.}}\\
    \midrule
    \textit{Rand} & 68.1 & 52.1 & 37.4 & 60.3 & 64.2 & 61.5 & 39.0 & 26.8 & 21.6 & 62.2 & 49.3 & 55.9 & 53.1 & 32.7 & 60.2 & 62.9 & 62.6 & 36.3 & 26.6 & 19.5 & 46.2 & 45.6 & 53.6 & 52.4 & 34.3 & 59.2 & 59.7 & 62.7 & 38.4 & 25.6 & 16.2 & 39.4 & 44.1 \\
    \midrule
    \multicolumn{34}{c}{\textbf{\textit{Feature-based Baselines}}} \\
    \textit{KMC*} & 58.3 & 52.5 & 23.9 & 59.0 & 64.5 & 61.5 & 41.1 & 26.5 & 15.3 & 35.9 & 43.9 & 48.3 & 52.3 & 37.0 & 59.0 & 62.0 & 62.6 & 31.6 & 23.4 & 22.9 & 37.1 & 43.6 & 63.7 & 51.2 & 40.8 & 60.6 & 58.8 & 61.7 & 35.6 & 26.7 & 19.4 & 49.6 & 46.8 \\
    \textit{MTLD*} & 64.7 & 52.4 & 35.2 & 61.0 & 62.5 & 62.6 & 43.2 & 18.3 & 7.2 & 57.7 & 46.5 & 67.3 & 48.6 & 29.1 & 58.7 & 64.6 & 62.7 & 31.9 & 23.1 & 13.6 & 61.0 & 46.1 & 65.2 & 51.1 & 23.8 & 58.1 & 61.4 & 60.5 & 34.2 & 20.5 & 3.6 & 58.1 & 43.7 \\
\bottomrule
    \end{tabular}
}
\caption{The detailed performance results of different language subset on \textit{MMMLU} dataset.}
\label{tab:main_mmmlu}
\end{table*}

%% file: tables/main_mkqa.tex
\begin{table*}[p]
\centering
\setlength{\aboverulesep}{1pt} 
\setlength{\belowrulesep}{1pt} 
\renewcommand{\arraystretch}{1.1} 
\setlength{\tabcolsep}{1.5pt}

\resizebox{\linewidth}{!}{
    \begin{tabular}{lcccccccccc|c|cccccccccc|c|cccccccccc|c}
    \toprule
    \multirow{3}[3]{*}{\textbf{Method}} & \multicolumn{33}{c}{\textbf{\textit{Model: LLaMA-3.1-8B}}} \\
    \cmidrule(lr){2-34}
    & \multicolumn{33}{c}{\textbf{Dataset: MKQA}} \\
    \cmidrule(lr){2-34}
     & \multicolumn{11}{c}{1\%} & \multicolumn{11}{c}{3\%} & \multicolumn{11}{c}{5\%} \\
    \midrule
    Langs & en & ar & de & ja & ko & pt & ru & tr & vi & zh & {\textbf{Avg.}} & en & ar & de & ja & ko & pt & ru & tr & vi & zh & {\textbf{Avg.}} & en & ar & de & ja & ko & pt & ru & tr & vi & zh & {\textbf{Avg.}}\\
    \midrule
    \textit{Rand} & 31.2 & 7.0 & 21.4 & 10.0 & 6.1 & 18.5 & 10.8 & 16.2 & 16.6 & 8.6 & 14.6 & 31.9 & 5.7 & 22.7 & 8.8 & 5.4 & 19.3 & 8.9 & 15.9 & 16.4 & 9.1 & 14.4 & 30.9 & 6.0 & 22.1 & 8.3 & 5.0 & 20.2 & 9.9 & 15.7 & 17.2 & 9.1 & 14.4 \\
    \midrule
    \multicolumn{34}{c}{\textbf{\textit{Feature-based Baselines}}} \\
    
    \textit{KMC*} & 31.4 & 7.4 & 21.2 & 11.0 & 6.0 & 18.9 & 10.6 & 16.7 & 19.0 & 8.6 & 15.1 & 33.5 & 5.9 & 22.8 & 8.0 & 5.4 & 20.6 & 9.3 & 15.5 & 17.2 & 9.7 & 14.8 & 32.0 & 6.6 & 22.2 & 8.2 & 6.0 & 20.8 & 9.9 & 16.7 & 17.2 & 10.4 & 15.0 \\
    \textit{MTLD} & 30.2 & 8.5 & 21.6 & 11.4 & 6.5 & 20.6 & 11.0 & 18.7 & 18.9 & 9.6 & 15.7 & 33.6 & 8.1 & 22.2 & 11.7 & 7.0 & 21.4 & 9.1 & 17.7 & 17.8 & 10.2 & 15.9 & 29.1 & 7.8 & 19.2 & 10.4 & 7.7 & 18.2 & 9.6 & 18.6 & 16.1 & 8.8 & 14.5 \\
\bottomrule
    \end{tabular}
}
\resizebox{\linewidth}{!}{
    \begin{tabular}{lcccccccccc|c|cccccccccc|c|cccccccccc|c}
    \toprule
    \multirow{3}[3]{*}{\textbf{Method}} & \multicolumn{33}{c}{\textbf{\textit{Model: Qwen2.5-7B}}} \\
    \cmidrule(lr){2-34}
    & \multicolumn{33}{c}{\textbf{Dataset: MKQA}} \\
    \cmidrule(lr){2-34}
     & \multicolumn{11}{c}{1\%} & \multicolumn{11}{c}{3\%} & \multicolumn{11}{c}{5\%} \\
    \midrule
    Langs & en & ar & de & ja & ko & pt & ru & tr & vi & zh & {\textbf{Avg.}} & en & ar & de & ja & ko & pt & ru & tr & vi & zh & {\textbf{Avg.}} & en & ar & de & ja & ko & pt & ru & tr & vi & zh & {\textbf{Avg.}} \\
    \midrule
    \textit{Rand} & 31.9 & 6.9 & 19.4 & 11.9 & 8.3 & 21.6 & 10.2 & 12.8 & 16.4 & 15.7 & 15.5 & 30.5 & 6.7 & 19.3 & 11.9 & 7.9 & 20.0 & 9.5 & 13.1 & 16.6 & 15.2 & 15.1 & 28.6 & 6.5 & 19.0 & 10.8 & 7.7 & 20.4 & 9.0 & 13.5 & 15.7 & 14.3 & 14.5 \\
    \midrule
    \multicolumn{34}{c}{\textbf{\textit{Feature-based Baselines}}} \\
    \textit{KMC*} & 28.4 & 6.1 & 17.0 & 9.7 & 7.2 & 17.6 & 6.6 & 9.8 & 14.6 & 14.2 & 13.1 & 29.4 & 6.0 & 17.1 & 9.1 & 7.4 & 20.2 & 10.4 & 12.4 & 15.3 & 14.7 & 14.2 & 28.2 & 5.5 & 16.3 & 10.1 & 6.7 & 19.1 & 7.9 & 11.8 & 14.5 & 13.8 & 13.4 \\
    \textit{MTLD*} & 32.3 & 7.4 & 19.0 & 12.4 & 8.1 & 20.9 & 9.7 & 12.0 & 15.2 & 15.2 & 15.2 & 29.7 & 6.7 & 17.6 & 12.6 & 7.7 & 18.7 & 8.5 & 12.2 & 13.6 & 15.2 & 14.2 & 29.3 & 7.3 & 18.5 & 11.8 & 7.7 & 20.1 & 9.0 & 12.7 & 14.8 & 14.6 & 14.6 \\
\bottomrule
    \end{tabular}
}
\caption{The detailed performance results of different language subset on \textit{MKQA} dataset.}
\label{tab:main_mkqa}
\end{table*}

%% file: tables/main_xquad.tex
\begin{table*}[p]
\centering
\setlength{\aboverulesep}{1pt} 
\setlength{\belowrulesep}{1pt} 
\renewcommand{\arraystretch}{1.1} 
\setlength{\tabcolsep}{1.5pt}

\resizebox{\linewidth}{!}{
    \begin{tabular}{lcccccccccc|c|cccccccccc|c|cccccccccc|c}
    \toprule
    \multirow{3}[3]{*}{\textbf{Method}} & \multicolumn{33}{c}{\textbf{\textit{Model: LLaMA-3.1-8B}}} \\
    \cmidrule(lr){2-34}
    & \multicolumn{33}{c}{\textbf{Dataset: XQuAD}} \\
    \cmidrule(lr){2-34}
     & \multicolumn{11}{c}{1\%} & \multicolumn{11}{c}{3\%} & \multicolumn{11}{c}{5\%} \\
    \midrule
    Langs & en & ar & de & el & es & hi & ru & th & tr & zh & {\textbf{Avg.}} & en & ar & de & el & es & hi & ru & th & tr & zh & {\textbf{Avg.}} & en & ar & de & el & es & hi & ru & th & tr & zh & {\textbf{Avg.}} \\
    \midrule
    \textit{Rand} & 75.1 & 58.4 & 67.6 & 53.9 & 68.3 & 62.5 & 53.8 & 63.7 & 58.0 & 74.9 & 63.6 & 73.0 & 52.9 & 64.5 & 53.2 & 66.6 & 59.2 & 50.1 & 55.5 & 52.4 & 72.2 & 59.9 & 69.7 & 53.9 & 63.8 & 50.9 & 64.5 & 60.9 & 48.6 & 60.4 & 51.8 & 71.9 & 59.6 \\
    \midrule
    \multicolumn{34}{c}{\textbf{\textit{Feature-based Baselines}}} \\
    
    \textit{KMC*} & 71.6 & 54.2 & 63.1 & 51.8 & 60.9 & 62.9 & 48.7 & 61.7 & 48.5 & 73.3 & 59.7 & 72.7 & 52.9 & 61.9 & 50.0 & 65.1 & 61.8 & 47.8 & 56.6 & 49.6 & 70.3 & 58.9 & 72.7 & 53.6 & 61.9 & 50.8 & 66.5 & 60.7 & 51.0 & 57.7 & 47.4 & 71.0 & 59.3 \\
    \textit{MTLD} & 74.3 & 55.5 & 64.8 & 52.5 & 62.7 & 60.4 & 50.5 & 59.2 & 51.8 & 71.4 & 60.3 & 78.7 & 55.5 & 65.7 & 54.9 & 66.1 & 63.5 & 53.4 & 59.6 & 51.9 & 75.5 & 62.5 & 65.3 & 52.9 & 61.8 & 50.8 & 63.8 & 59.7 & 47.7 & 56.0 & 46.4 & 67.6 & 57.2 \\
\bottomrule
    \end{tabular}
}
\resizebox{\linewidth}{!}{
    \begin{tabular}{lcccccccccc|c|cccccccccc|c|cccccccccc|c}
    \toprule
    \multirow{3}[3]{*}{\textbf{Method}} & \multicolumn{33}{c}{\textbf{\textit{Model: Qwen2.5-7B}}} \\
    \cmidrule(lr){2-34}
    & \multicolumn{33}{c}{\textbf{Dataset: XQuAD}} \\
    \cmidrule(lr){2-34}
     & \multicolumn{11}{c}{1\%} & \multicolumn{11}{c}{3\%} & \multicolumn{11}{c}{5\%} \\
    \midrule
    Langs & en & ar & de & el & es & hi & ru & th & tr & zh & {\textbf{Avg.}} & en & ar & de & el & es & hi & ru & th & tr & zh & {\textbf{Avg.}} & en & ar & de & el & es & hi & ru & th & tr & zh & {\textbf{Avg.}} \\
    \midrule
    \textit{Rand} & 77.6 & 65.4 & 69.9 & 46.0 & 71.2 & 54.4 & 52.7 & 74.4 & 58.9 & 83.7 & 65.4 & 77.5 & 64.9 & 70.3 & 42.1 & 72.2 & 47.7 & 53.8 & 69.9 & 57.7 & 83.5 & 64.0 & 78.3 & 64.8 & 70.4 & 40.6 & 73.9 & 47.5 & 54.4 & 67.1 & 58.4 & 83.7 & 63.9 \\
    \midrule
    \multicolumn{34}{c}{\textbf{\textit{Feature-based Baselines}}} \\
    \textit{KMC*} & 73.7 & 59.8 & 68.0 & 38.9 & 66.5 & 52.3 & 47.1 & 70.8 & 54.0 & 81.3 & 61.2 & 78.1 & 66.1 & 68.6 & 36.1 & 72.0 & 44.2 & 52.3 & 65.9 & 53.3 & 83.9 & 62.0 & 78.1 & 63.4 & 71.1 & 39.4 & 71.3 & 44.4 & 52.1 & 65.3 & 56.1 & 82.9 & 62.4 \\
    \textit{MTLD*} & 78.6 & 65.0 & 71.5 & 44.5 & 72.1 & 54.1 & 53.7 & 73.2 & 56.0 & 86.8 & 65.5 & 78.6 & 61.3 & 70.0 & 40.8 & 71.4 & 51.6 & 55.9 & 67.2 & 53.2 & 85.2 & 63.5 & 79.5 & 62.1 & 70.4 & 38.0 & 73.3 & 46.8 & 53.5 & 65.3 & 53.9 & 84.8 & 62.8 \\
\bottomrule
    \end{tabular}
}
\caption{The detailed performance results of different language subset on \textit{XQuAD} dataset.}
\label{tab:main_xquad}
\end{table*}

%% file: tables/main_xlsum.tex
\begin{table*}[p]
\centering
\setlength{\aboverulesep}{1pt} 
\setlength{\belowrulesep}{1pt} 
\renewcommand{\arraystretch}{1.1} 
\setlength{\tabcolsep}{1.5pt}

\resizebox{\linewidth}{!}{
    \begin{tabular}{lcccccccccc|c|cccccccccc|c|cccccccccc|c}
    \toprule
    \multirow{3}[3]{*}{\textbf{Method}} & \multicolumn{33}{c}{\textbf{\textit{Model: LLaMA-3.1-8B}}} \\
    \cmidrule(lr){2-34}
    & \multicolumn{33}{c}{\textbf{Dataset: XLSum}} \\
    \cmidrule(lr){2-34}
     & \multicolumn{11}{c}{1\%} & \multicolumn{11}{c}{3\%} & \multicolumn{11}{c}{5\%} \\
    \midrule
    Langs & en & es & fr & ko & pt & sw & tr & uk & vi & zh & {\textbf{Avg.}} & en & es & fr & ko & pt & sw & tr & uk & vi & zh & {\textbf{Avg.}} & en & es & fr & ko & pt & sw & tr & uk & vi & zh & {\textbf{Avg.}} \\
    \midrule
    \textit{Rand} & 26.1 & 21.8 & 27.3 & 14.5 & 24.2 & 22.6 & 20.2 & 17.5 & 18.9 & 25.9 & 21.9 & 29.3 & 21.6 & 28.9 & 14.9 & 25.2 & 20.8 & 21.8 & 17.4 & 20.6 & 27.9 & 22.8 & 28.3 & 21.3 & 27.3 & 14.4 & 24.8 & 20.8 & 22.8 & 17.3 & 22.0 & 28.9 & 22.8 \\
    \midrule
    \multicolumn{34}{c}{\textbf{\textit{Feature-based Baselines}}} \\
    
    \textit{KMC*} & 26.3 & 20.1 & 24.7 & 15.3 & 23.0 & 18.7 & 19.7 & 17.2 & 19.2 & 25.8 & 21.0 & 29.3 & 22.0 & 27.8 & 16.1 & 25.5 & 21.2 & 22.4 & 15.8 & 22.1 & 29.1 & 23.1 & 27.1 & 21.5 & 28.2 & 13.8 & 23.4 & 22.0 & 23.1 & 17.0 & 20.6 & 27.7 & 22.4 \\
    \textit{MTLD} & 24.2 & 19.5 & 24.2 & 13.4 & 21.6 & 18.3 & 18.5 & 13.3 & 16.4 & 25.2 & 19.5 & 26.5 & 20.3 & 27.6 & 15.2 & 23.9 & 21.3 & 21.6 & 16.3 & 20.2 & 26.3 & 21.9 & 33.0 & 20.8 & 29.5 & 15.2 & 26.4 & 22.7 & 23.4 & 19.2 & 23.9 & 27.6 & 24.2 \\
\bottomrule
    \end{tabular}
}
\resizebox{\linewidth}{!}{
    \begin{tabular}{lcccccccccc|c|cccccccccc|c|cccccccccc|c}
    \toprule
    \multirow{3}[3]{*}{\textbf{Method}} & \multicolumn{33}{c}{\textbf{\textit{Model: Qwen2.5-7B}}} \\
    \cmidrule(lr){2-34}
    & \multicolumn{33}{c}{\textbf{Dataset: XLSum}} \\
    \cmidrule(lr){2-34}
     & \multicolumn{11}{c}{1\%} & \multicolumn{11}{c}{3\%} & \multicolumn{11}{c}{5\%} \\
    \midrule
    Langs & en & es & fr & ko & pt & sw & tr & uk & vi & zh & {\textbf{Avg.}} & en & es & fr & ko & pt & sw & tr & uk & vi & zh & {\textbf{Avg.}} & en & es & fr & ko & pt & sw & tr & uk & vi & zh & {\textbf{Avg.}} \\
    \midrule
    \textit{Rand} & 25.9 & 21.9 & 28.4 & 15.5 & 23.7 & 19.9 & 20.6 & 17.9 & 20.4 & 26.6 & 22.1 & 26.2 & 21.7 & 29.0 & 15.5 & 23.0 & 18.2 & 21.9 & 17.2 & 21.1 & 28.3 & 22.2 & 26.4 & 21.4 & 29.0 & 16.0 & 23.4 & 20.2 & 21.5 & 17.6 & 21.6 & 29.6 & 22.7 \\
    \midrule
    \multicolumn{34}{c}{\textbf{\textit{Feature-based Baselines}}} \\
    \textit{KMC*} & 27.4 & 20.7 & 29.6 & 16.9 & 24.3 & 20.3 & 21.2 & 15.8 & 19.7 & 29.1 & 22.5 & 25.9 & 22.6 & 26.5 & 15.6 & 24.9 & 18.0 & 22.5 & 17.4 & 20.3 & 26.9 & 22.0 & 26.8 & 21.4 & 26.0 & 14.1 & 23.3 & 21.3 & 19.3 & 15.8 & 20.8 & 29.4 & 21.8 \\
    \textit{MTLD*} & 25.7 & 21.8 & 27.8 & 15.0 & 21.0 & 19.7 & 21.2 & 16.7 & 21.2 & 26.7 & 21.7 & 29.6 & 20.4 & 27.7 & 17.3 & 23.0 & 19.1 & 21.9 & 16.0 & 22.0 & 29.2 & 22.6 & 28.9 & 21.7 & 28.3 & 16.1 & 23.4 & 16.2 & 21.0 & 17.9 & 21.3 & 29.4 & 22.4 \\
\bottomrule
    \end{tabular}
}
\caption{The detailed performance results of different language subset on \textit{XLSum} dataset.}
\label{tab:main_xlsum}
\end{table*}